\documentclass[journal, one-side, wed]{cas-sc}
\usepackage[]{natbib}
\usepackage{threeparttable}
\usepackage[]{natbib}
\usepackage{amssymb}
\usepackage{amsmath}
\usepackage{diagbox}
\usepackage{subfigure}
\usepackage{graphicx}
\usepackage{verbatim}
\usepackage{siunitx}
\usepackage{booktabs}
\usepackage{multirow}
\usepackage{cases}
\usepackage{makecell}
\usepackage{pifont} 
\def\tsc#1{\csdef{#1}{\textsc{\lowercase{#1}}\xspace}}
\tsc{WGM}
\tsc{QE}
\tsc{EP}
\tsc{PMS}
\tsc{BEC}
\tsc{DE}

\begin{document}

\let\WriteBookmarks\relax
\def\floatpagepagefraction{1}
\def\textpagefraction{.001}

\title [mode = title]{Deep Learning-Based Object Detection in Maritime Unmanned Aerial Vehicle Imagery: Review and Experimental Comparisons}

\shorttitle{Deep Learning-Based Object Detection}

\author[1,2,3]{Chenjie Zhao} 
\ead{cjzhao@whut.edu.cn}  
\shortauthors{C. Zhao, et al.}

\author[1,2,3]{Ryan Wen Liu} 
\cormark[1]
\ead{wenliu@whut.edu.cn}

\author[1,2]{Jingxiang Qu} 
\ead{qujx@whut.edu.cn}

\author[4]{Ruobin Gao}
\ead{gaor0009@e.ntu.edu.sg}

\address[1]{Hubei Key Laboratory of Inland Shipping Technology, School of Navigation, Wuhan University of Technology, Wuhan 430063, China}

\address[2]{State Key Laboratory of Maritime Technology and Safety, Wuhan 430063, China}

\address[3]{Chongqing Research Institute, Wuhan University of
Technology, Chongqing, China}

\address[4]{School of Civil and Environmental Engineering, Nanyang Technological University, Singapore}

\cortext[cor1]{Corresponding author}

\begin{abstract}
    With the advancement of maritime unmanned aerial vehicles (UAVs) and deep learning technologies, the application of UAV-based object detection has become increasingly significant in the fields of maritime industry and ocean engineering. Endowed with intelligent sensing capabilities, the maritime UAVs enable effective and efficient maritime surveillance. 
    To further promote the development of maritime UAV-based object detection, this paper provides a comprehensive review of challenges, relative methods, and UAV aerial datasets. Specifically, in this work, we first briefly summarize four challenges for object detection on maritime UAVs, i.e., object feature diversity, device limitation, maritime environment variability, and dataset scarcity. We then focus on computational methods to improve maritime UAV-based object detection performance in terms of scale-aware, small object detection, view-aware, rotated object detection, lightweight methods, and others. Next, we review the UAV aerial image/video datasets and propose a maritime UAV aerial dataset named MS2ship for ship detection. Furthermore, we conduct a series of experiments to present the performance evaluation and robustness analysis of object detection methods on maritime datasets. Eventually, we give the discussion and outlook on future works for maritime UAV-based object detection. The MS2ship dataset is available at \href{https://github.com/zcj234/MS2ship}{https://github.com/zcj234/MS2ship}.
\end{abstract}

\maketitle

\section{Introduction} 
\begin{figure}[t]
    	\centering
    	\includegraphics[width=0.55 \textwidth]{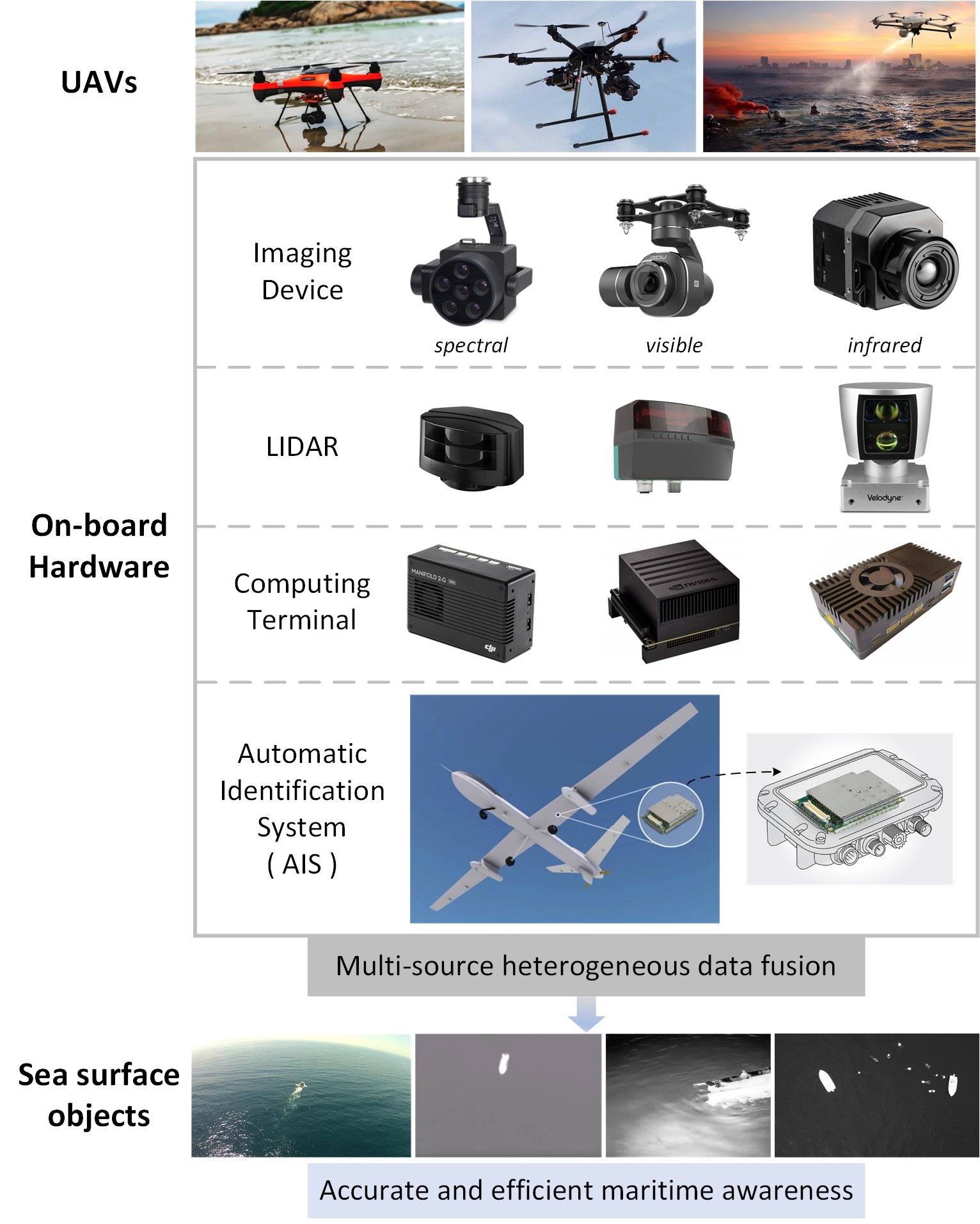}
    	\caption{The diagram of the UAV platform, on-board hardware, and  observed objects. Through multi-source heterogeneous data fusion, the maritime UAVs will achieve accurate and efficient maritime awareness.} 
    	\label{UAV_introduction_label} 
    \end{figure}

    The unmanned aerial vehicles (UAVs) have been used in various applications under dull, hazardous, or concealed conditions, such as search and rescue \citep{yang2020maritime}, traffic analyzing \citep{benjdira2022tau}, and emergency disaster response \citep{adams2011survey}. Owing to cutting-edge technologies, especially artificial intelligence (AI) and computer vision (CV), UAVs can now perform a wide range of tasks more efficiently and intelligently. In the fields of maritime industry and ocean engineering, successful applications have demonstrated the vast potential of maritime UAVs.
    The maritime UAVs are highly mobile and capable of executing complex maritime tasks without being limited by geographical environments. Additionally, they can carry various sensors, including cameras, light detection and ranging (LIDAR), and airborne Automatic Identification Systems (AIS), etc., to provide diversified data sources for maritime surveillance and rescue tasks. The functions of maritime UAVs ultimately depend on the sensors deployed on them \citep{gao2014study}. From Fig. \ref{UAV_introduction_label}, the UAV platform loading specific on-board hardware is able to achieve accurate and efficient maritime awareness by the multi-source heterogeneous data fusion technology. For instance, it can observe sea surface objects with cameras, and the infrared camera is particularly useful in low-light environments. Meanwhile, the AIS is able to transmit navigation information back to the control center, helping to monitor blind waters where communication facilities are damaged \citep{duan2014research}. With an on-board computing terminal and efficient models, the maritime UAV achieves data processing in real-time. 
    \cite{muhammad2022maritime} introduced maritime UAV services from an ecosystem perspective, providing an overview of potential applications of UAVs in the field. These applications primarily include daily patrols, emergency investigation/collection, and emergency rescue of marine accidents. Besides, the maritime UAV could serve as an essential supplement to maritime law enforcement, emergency response, communication relay, and beacon inspection \citep{nomikos2022survey}. Therefore, it is undeniable to affirm the vast potential of maritime UAVs in maritime industry and ocean engineering.

    Recently, object detection in UAV aerial images has gained great momentum, especially in the maritime industry. The applications, such as surveillance and rescue, require efficient object detection in drone views. As one of the crucial tasks in the CV field, object detection is applied to determine the classifications and positions of objects. Traditional object detection methods are based on handcrafted features. In 2001, Viola Jones detector was proposed to detect human face \citep{viola2001rapid}. However, the calculation of their sliding window method was far beyond the computer power in that time. The histogram of oriented gradients (HOG) detector \citep{dalal2005histograms} was proposed in 2005 to detect pedestrian, which had been the foundation of subsequent detectors for many years. The deformable part-based model (DPM) \citep{felzenszwalb2008discriminatively} followed the divide and conquer strategy. Roughly stated, the training is considered as the learning of decomposing an object, while the inference is regarded as an ensemble of detections on different object parts. In general, traditional methods based on sliding window have high time complexity due to lots of redundant windows, as well as the low power of computational devices. With the rapid development of graphics processing unit (GPU), a strong parallel computing device, deep learning technologies have attracted extensive attention in the CV field. At the same time, deep learning-based object detection has also achieved remarkable success in both detection accuracy and efficiency, whose wide applications have been witnessed in multiple industrial domains, e.g., self-driving \citep{gupta2021deep}, underwater navigation \citep{9791304}, and intelligent maritime surveillance \citep{qu2022intelligent}.

    The object detection technique has promoted the development of intelligent visual systems, such as the UAV platform with visual functions. Moreover, object detection in UAV aerial images has also attracted increasing attention and brought innovative solutions to maritime industry problems \citep{kong2022object}. The UAV-based object detection in the maritime domain is identified as an important aspect requiring further research and development. When equipped with both cameras and computing devices, the maritime UAVs have the ability to detect and track various sea surface objects, such as ships, offshore facilities, and marine pollutants, which greatly improves the efficiency of maritime surveillance and helps to protect maritime environments. Moreover, another crucial application is related to search and rescue missions. By using infrared cameras, the maritime UAVs are capable of working at night and detecting drowning persons \citep{feraru2020towards}. As researchers continue to develop advanced object detection techniques, the capability of maritime UAVs will undoubtedly continue to expand in the field of maritime industry. 

\begin{table}[] 
    \begin{center}
        \caption{Summary of recent surveys related to UAV-based object detection based on deep learning technologies.}
\renewcommand\arraystretch{1.2}
\footnotesize
\resizebox{\linewidth}{!}{
\begin{tabular}{llll}
\hline
Title                                           & Descriptions                                                                                                                                                                                                                                                                                                                                                                  & Publisher                                                                                               & Year \\ \hline
\cite{cazzato2020survey}      & \begin{tabular}[c]{@{}l@{}}Presents the latest advances in UAV-based object detection and focuses \\ on the differences, strategies, and trade-offs between the generic problem \\ of object detection and their adaptability on UAVs.\end{tabular}                                                                                                   & Journal of Imaging                                                                                      & 2020 \\ \hline
\cite{mittal2020deep}         & \begin{tabular}[c]{@{}l@{}}Provides a comprehensive review of state-of-the-art deep learning-based \\ object detection algorithms and analyses recent contributions of these \\ algorithms to low-altitude UAV datasets.\end{tabular}                                                                                                                                     & \begin{tabular}[c]{@{}l@{}}Image and Vision \\ Computing\end{tabular}                                   & 2020 \\ \hline
\cite{srivastava2021survey}   & \begin{tabular}[c]{@{}l@{}}Reviews deep learning techniques for performing on-ground vehicle \\ detection from UAV aerial images and efforts to improve accuracy and \\ reduce computational overhead, as well as their optimization objective.\end{tabular}                                                                                             & \begin{tabular}[c]{@{}l@{}}Journal of Systems \\ Architecture\end{tabular}                              & 2021 \\ \hline
\cite{bouguettaya2021vehicle} & \begin{tabular}[c]{@{}l@{}}Focuses on the vehicle detection task, and outlines various deep learning \\ techniques for improving vehicle detection in UAV aerial images.\end{tabular}                                                                                                                                                                                              & \begin{tabular}[c]{@{}l@{}}IEEE Transactions on \\ Neural Networks and \\ Learning Systems\end{tabular} & 2021 \\ \hline
\cite{zhang2021survey}        & \begin{tabular}[c]{@{}l@{}}Summarizes methods and application scenarios of maritime object detection, \\ especially YOLO series, and also discusses current limitations of deep \\ learning-based object detection and possible breakthrough directions.\end{tabular}                                         & \begin{tabular}[c]{@{}l@{}}Journal of Advanced \\ Transportation\end{tabular}                           & 2021 \\ \hline
\cite{nguyen2022state}        & \begin{tabular}[c]{@{}l@{}}A comprehensive overview of human-centered aerial surveillance tasks from \\ computer vision and pattern recognition perspectives, and their current \\ applications on drones, UAVs, and other airborne platforms.\end{tabular}              & ArXiv                                                                                                   & 2022 \\ \hline
\cite{9604009}                & \begin{tabular}[c]{@{}l@{}}A comprehensive review of the research progress and prospects of deep \\ learning-based UAV object detection and tracking methods, including \\ challenges and statistics of existing methods.\end{tabular}  & \begin{tabular}[c]{@{}l@{}}IEEE Geoscience \\ and Remote \\ Sensing Magazine\end{tabular}               & 2022 \\ \hline
\cite{rekavandi2022guide}     & \begin{tabular}[c]{@{}l@{}}Explores deep learning-based small object detection (SOD) on both \\ images and videos, and summarizes datasets for SOD in general and \\ maritime applications.\end{tabular}                                                                                                                                     & ArXiv                                                                                                   & 2022 \\ \hline
\cite{lyu2022sea}             & \begin{tabular}[c]{@{}l@{}}Provides a comprehensive overview of maritime object detection methods, \\ and compares the advantages and disadvantages of each technique from \\ four fundamental aspects, i.e., electro-optical sensors, traditional object \\ detection methods, deep learning methods, and maritime datasets.\end{tabular}        & \begin{tabular}[c]{@{}l@{}}IEEE Intelligent \\ Transportation \\ Systems Magazine\end{tabular}          & 2022 \\ \hline
Our work                                      & \begin{tabular}[c]{@{}l@{}}For the object detection tasks on maritime UAVs, discusses major \\ challenges and reviews the corresponding methods based on deep \\ learning. And a UAV aerial dataset and experiments for maritime object \\ detection are provided.\end{tabular}                                                       & \begin{tabular}[c]{@{}l@{}}Engineering Applications \\ of Artificial Intelligence\end{tabular}                                                                                       & 2023 \\ \hline
\end{tabular}
}
\end{center}
\label{survey_table}
\end{table}
    Even if deep learning-based object detection has reached a mature stage, it is still unsatisfactory when it comes to detecting maritime UAV aerial images, because of several challenging factors, such as the complexity of ocean environments, complex weather conditions, and the existence of small and low-contrast objects. Additionally, significant differences between UAV aerial images and natural scene images present a challenge in the direct utilization of object detection models learned from general datasets. Despite the growing interest in this area, there are only several published reviews on object detection tasks for UAV aerial images, especially in the maritime domain. 
    Table \ref{survey_table} summarizes the critical surveys related to UAV-based object detection based on deep learning technologies in recent years. However, these reviews have not provided holistic and general overviews of the state-of-the-art UAV-based object detection for maritime visual tasks.

    The challenges and unique features of maritime UAV-based object detection, e.g., complex environments, adverse weather, and small objects, bring strict requirements to the detection methods. Therefore, in this study, we provide a holistic review of deep learning-based object detection, specifically for the application of maritime UAVs, covering not only methods from different aspects but also UAV-based datasets. Moreover, adequate experiments about representative methods are also conducted. The findings and insights of our survey and experimental analysis could help to guide the development of more efficient and accurate object detection methods on maritime UAVs. The main contributions of this paper are summarized as follows:
  
    \begin{itemize}
	\item Providing a detailed analysis of major challenges for maritime UAV-based object detection and reviewing corresponding deep learning-based methods, which can be grouped into six categories: scale-aware, small object detection, view-aware, rotated object detection, lightweight methods, and others.

	\item Reviewing the UAV aerial datasets that span over multiple scenarios and then proposing a ship image dataset, named MS2ship. We hope the annotated benchmark could help researchers to develop and verify ship detection methods appropriately for maritime UAV aerial images.

	\item Conducting a thorough evaluation of state-of-the-art object detection methods on maritime datasets, and providing a detailed discussion of experimental results. The analysis indicates the strengths and weaknesses of compared methods and their performance in maritime scenes.
    \end{itemize}

    The remainder of this paper is organized as follows: In Section 2, we discuss major challenges for maritime UAV-based object detection. Section 3 makes a thorough review of deep learning-based object detection methods, particularly for applications on maritime UAVs. In section 4, we investigate UAV-based datasets and propose a ship image dataset for ship detection task on maritime UAVs. Section 5 provides experimental results and analysis of representative methods on maritime datasets and then make a detailed discussion. Finally, we make a brief conclusion.

\section{Bibliometric Study}
    To explore characteristics of the research on deep learning-based object detection methods scientifically and comprehensively, we applied the big data analysis technique based on bibliometrics and studied the trend of publication volume, as well as the research hotspots. Differing from traditional reviews, the bibliometric analysis-based science mapping combines quantitative analysis, classification, and data visualisation, to provide a structured view of researched themes \citep{zupic2015bibliometric}. We searched the literature from the Web of Science (WOS), IEEE Xplore, and Scopus databases by the following indices (time thresholds 2015-2023).

    WOS: TS=(object detection) AND TS=(deep learning) AND (TS=(maritime) OR TS=(UAV) OR TS=(drone)) and Web of Science Core Collection (Database) and Computer Science or Engineering or Remote Sensing (Research Area) and Article or Proceeding Paper or Review Article or Early Access (Document Types). Total 411. Articles (338) and Other (173).

    SCOPUS: ABS (deep learning AND object detection AND (maritime OR UAV OR drone)) AND (LIMIT-TO (SRCTYPE, 'p') OR LIMIT-TO (SRCTYPE, 'j')) AND (LIMIT-TO (SUBJAREA, 'COMP')) AND (LIMIT-TO (DOCTYPE, 'cp') OR LIMIT-TO  (DOCTYPE, 'ar') OR LIMIT-TO (DOCTYPE, 're')) AND (LIMIT-TO (LANGUAGE, 'English')) AND (LIMIT-TO (EXACTKEYWORD, 'Deep Learning') OR LIMIT-TO (EXACTKEYWORD, 'Object Detection')) AND (LIMIT-TO (LANGUAGE, 'English')). Total 339. Article (110), Conference Paper (221), Review (8).

    IEEE Xplore: ('Abstract': deep learning) AND ('Abstract': object detection) AND ('Abstract': maritime OR 'Abstract': UAV OR 'Abstract': drone). Total 254. Conference (206), Journals and Magazines (44), Early Access Articles (4).
    
    \begin{figure}[t]
	\centering
	\begin{minipage}{1 \linewidth}
		\centering
		\includegraphics[width=0.5 \linewidth]{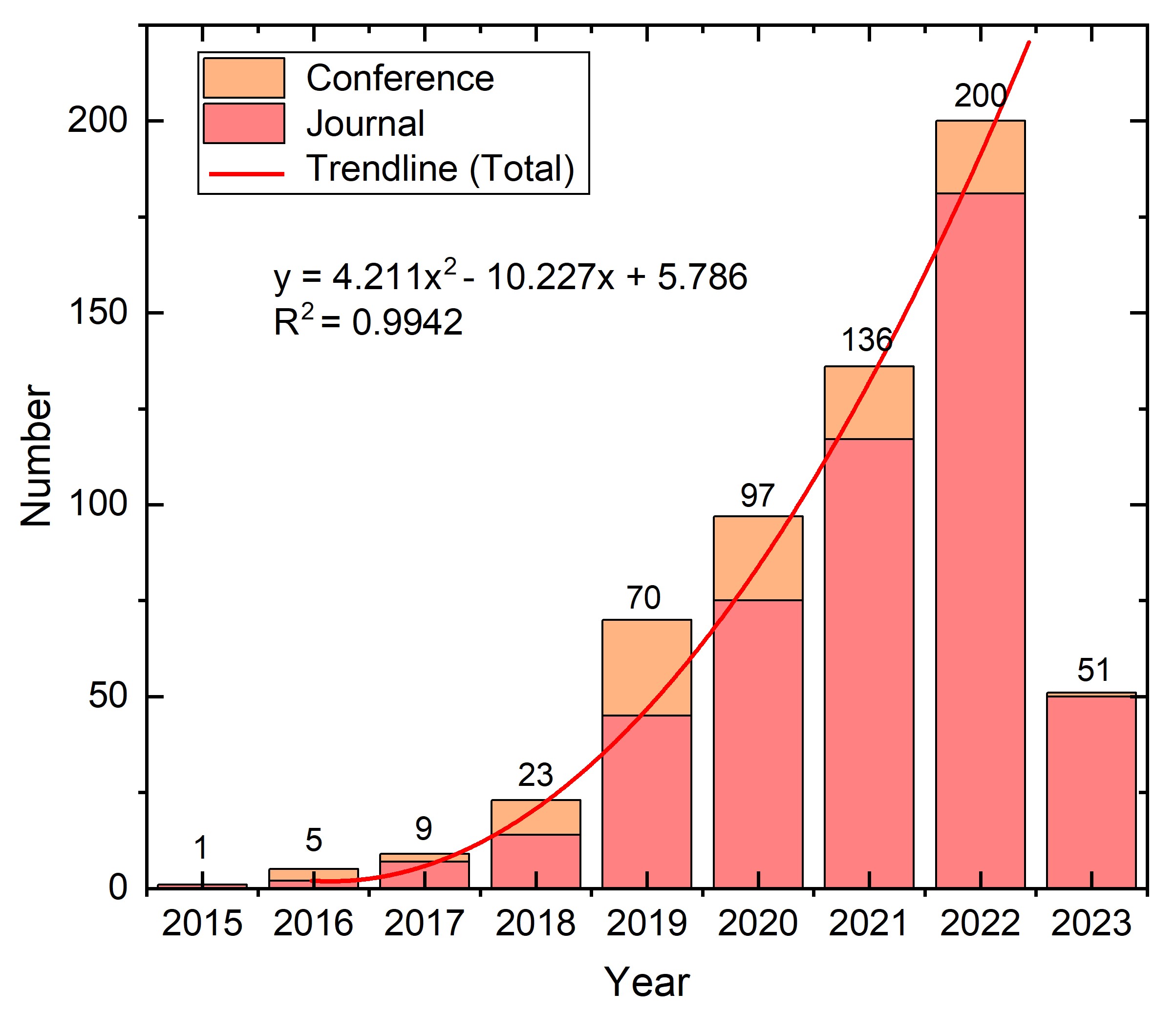}
		\caption{The chronological distribution of publications between 2015 and 2023. The data in 2023 are for the months from January 1st to May 1st.}
		\label{article_time}
    \end{minipage}
    \begin{minipage}{1 \linewidth}
		\centering
		\includegraphics[width=0.7 \linewidth]{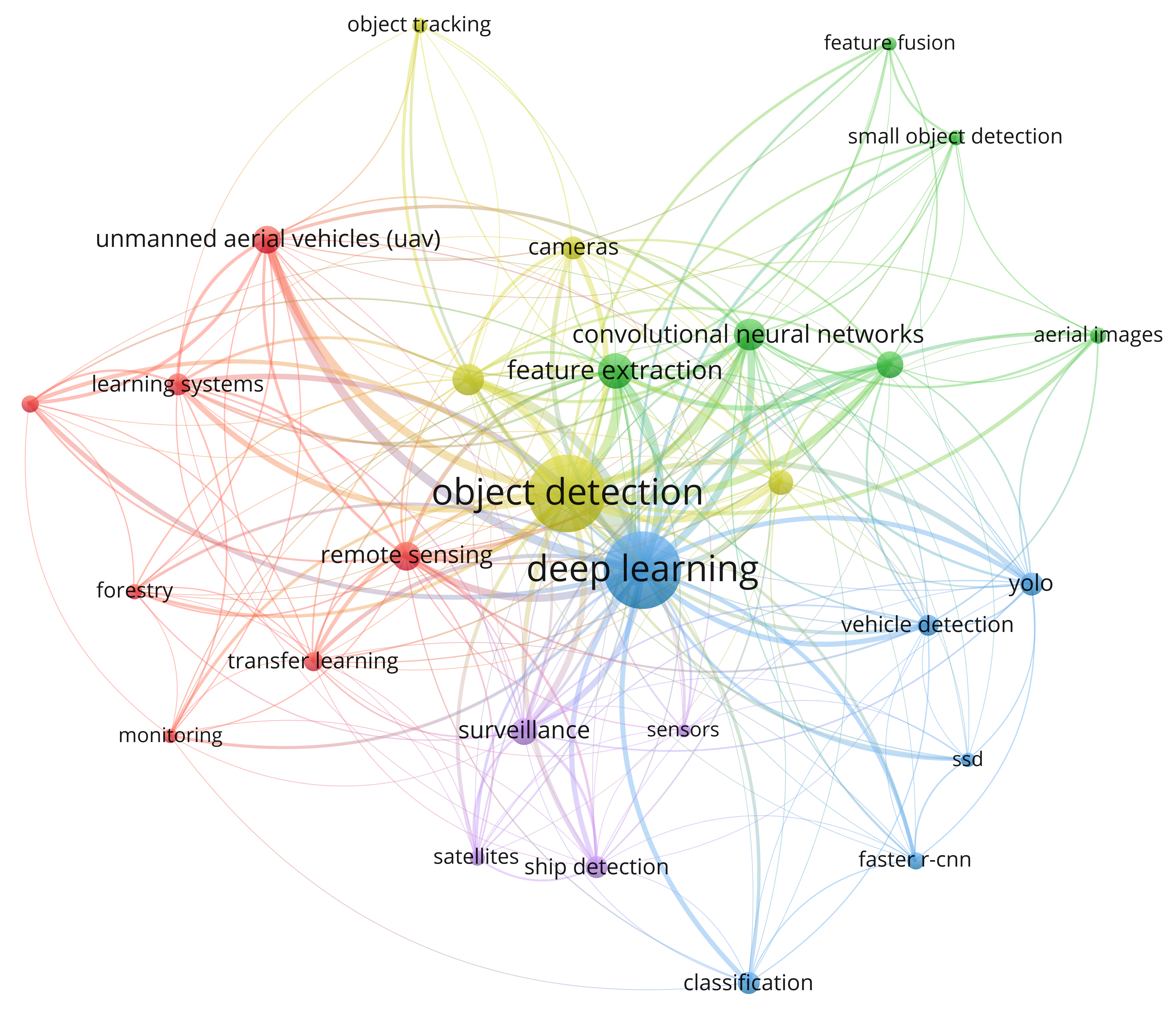}
		\caption{The co-occurring keyword-based analysis network. The size of each keyword node is proportional to the number of related literature. The thicker lines between nodes denote their stronger correlation.}
		\label{vosview_keywords}
	\end{minipage}
    \end{figure}

    Some of the searched literature study the detection of UAVs, but are unrelated to our survey contents. After eliminating these literature and some duplicated articles between different databases, a total of 592 items related to deep learning-based object detection in the UAV application and maritime domain were obtained. Fig. \ref{article_time} shows the chronological distribution of publications. Based on the annual issue volume, the correlation coefficient $R^2$ of the second-order polynomial trend line is 0.9942. The result indicates that object detection in specific fields is becoming more and more popular. The first article in 2015 \citep{tang2015compressed} explored ship detection by deep neural network, and since then, the number of articles doubled year by year. Furthermore, the growth mainly comes from journals while the number of conference papers just has a little change between 2019 and 2022. By May 2023, there had been already 51 articles, with more expected by the end of the year. Fig. \ref{vosview_keywords}, the network visualization of keyword co-occurrence analysis, highlights the hotspots of object detection in the UAV application and maritime domain. Specifically, a spot consists of the circle and label, whose size depends on the node degree, link strength, citations, and etc. The color of each element represents the corresponding cluster, and there are five colors, including red, yellow, blue, green, and purple. Through the bibliometric study, we are familiar with the study trend and research focus of relevant fields of maritime UAV-based object detection. Later, we will discuss its core content in detail.

\section{Challenges for Object Detection on Maritime UAVs} \label{challenge}
\begin{figure}[t]
	\centering
	\includegraphics[width=0.8 \linewidth]{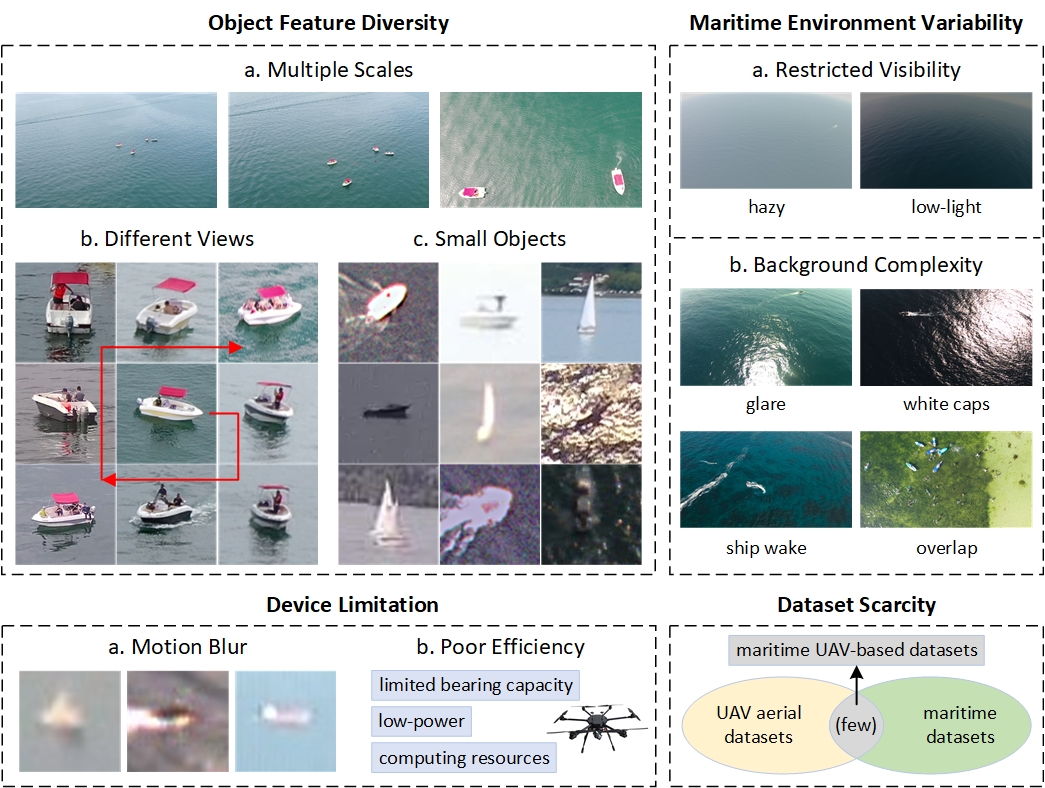}
	\caption{The four major challenges for maritime UAV-based object detection.}
	\label{challenges_fig} 
    \end{figure}

    The maritime UAV-based object detection is a challenging task, owing to different imaging conditions in the unpredictable and dynamic maritime environment. The processing for UAV-based images and videos can be divided into off-board and on-board. The off-board processing requires wireless data transmission, and the quality of visual data is often compromised by unstable communication, limited bandwidth, and transmission delay. In contrast, the on-board processing is more efficient and secure. Therefore, we mainly focus on the on-board methods for maritime UAV-based object detection. It poses unique challenges that need to be addressed, i.e., object feature diversity, device limitation, maritime environment variability, and dataset scarcity, as shown in Fig. \ref{challenges_fig}.

    \subsection{Object Feature Diversity}
        The maritime UAVs possess exceptional maneuverability, enabling them to move quickly and capture sea surface objects from various views. Consequently, object features in maritime images are diverse, encompassing multiple scales, small objects, and different views, which present significant challenges for maritime object detection.

        a. Multiple scales. 
            The maritime UAVs are able to capture images from varying altitudes and distances, leading to significant scale variation among observed objects. Even the captured objects of the same category may have different sizes. Therefore, it is ambiguous and confusing for the detection model to extract features.

        b. Small objects. 
            When maritime UAVs work at high altitudes with wide fields of view, the captured images always contain a mass of small objects. The indistinctive characteristics of small objects will increase mistaken detections and then decrease the detection accuracy. Additionally, small objects are prone to be disturbed by the maritime environment, requiring more precise positioning.

        c. Different views. 
            The mobility of maritime UAVs enables them the ability to capture objects from different views, including front, side, and bird's-eye views. However, various viewpoint changes lead to differences in the appearance of object features, resulting in the large intra-class difference and posing a significant challenge to the robustness of object detection models. 

    \subsection{Device Limitation}
        With the equipment of imaging and computing devices on maritime UAVs, advanced vision technologies such as object detection can be executed on board. However, the movement of maritime UAVs may cause motion blur, which leads to unqualified imaging data. Moreover, the computing power of on-board devices is always limited, leading to a decrease in efficiency. It is thus necessary to address these challenges for reliable and efficient object detection on maritime UAVs.

        a. Motion blur issue. 
            The motion blur is a common problem in object detection and tracking tasks when dealing with maritime UAV aerial images/videos. 
            The high-speed and low-altitude flight results in the motion blur on densely packed objects \citep{zhu2021tph}. In addition, a strong breeze will make the camera shake in the maritime environment. 
            The weak anti-wind capability of maritime UAVs also introduces blurs and additional noises to images. 

        b. Poor efficiency issue. 
            Detecting sea-surface objects quickly is essential for real-time tasks on maritime UAVs. However, the bearing capacity, power consumption, and computing resources carried by maritime UAVs are often limited, which makes efficiency poor and even causes failed detection. Therefore, it is vital to develop fast and accurate object detection methods that can be processed on board with limited computing resources.

    \subsection{Maritime Environment Variability}
        a. Restricted visibility. 
            Due to the complexity and variability of the maritime environment, adverse weather conditions, such as low light and haze, significantly reduce visibility, making it difficult to detect objects accurately. Therefore, it is crucial to develop object detection methods that can achieve relatively robust predictions under changeable illuminations and weather conditions.

        b. Background complexity. 
            The background complexity is another challenge presented by the maritime environment. Due to many interference factors, the background of maritime UAV aerial images is often cluttered. For instance, images and videos often suffer from the severe glare caused by the sunlight reflection. The water areas have unique elements, such as white caps caused by waves, also reducing the visibility of objects. Additionally, ships sailing in the sea always produce a wake, and there are always overlaps between human and boat. All these factors make it difficult to achieve accurate and reliable object detection on maritime UAVs.

    \subsection{Dataset Scarcity}
        The deep learning-based object detection requires image training data to achieve high accuracy and robustness in practical applications. However, the scarcity of available datasets is another challenge in developing maritime detection methods for UAV purposes. Most of the existing UAV aerial datasets, such as VisDrone \citep{zhu2018vision, zhu2020vision}, AU-AIR \citep{bozcan2020air}, and Stanford UAV dataset \citep{robicquet2016learning}, focus on inland scenes. The maritime datasets, such as SeaShips \citep{shao2018seaships}, Mcships \citep{zheng2020mcships}, and Singapore Maritime Dataset \citep{prasad2017video}, are mainly captured by shore-based, ship-borne or satellite-based platforms, which are different from the perspective of maritime UAV aerial dataset.

        Therefore, to overcome the challenge, it is essential to collect and annotate a qualified dataset that can serve for ship detection task on maritime UAVs. The dataset should include various ship types and a wide range of maritime scenes under different lighting and weather conditions, enabling the development of robust and accurate object detection models for maritime UAVs.

\section{Deep Learning-Based Object Detection}
    Throughout this section, we provide a comprehensive review of deep learning-based object detection methods. Firstly, a short survey on generic object detection methods is presented. However, they may encounter poor accuracy and efficiency due to the unique challenges in the maritime environments. Therefore, we present relevant methods aiming at tackling these challenges in Section \ref{3.2}.
    
    \subsection{Generic Methods}
        As deep learning technologies continue to advance, object detection is being applied in various fields, including vehicle detection \citep{mittal2023ensemblenet}, face detection \citep{liu2023center}, vessel detection \citep{guo2023thfe}, etc. Most deep learning-based object detection methods are based on convolutional neural network (CNN) \citep{krizhevsky2017imagenet} architectures, which are exploited to extract features from input images. Generally, they can be categorized into two-stage and one-stage detectors.

        Typical methods for two-stage detectors are based on regions with CNN feature (R-CNN) \citep{girshick2014rich}, which include Fast R-CNN \citep{girshick2015fast}, Faster R-CNN \citep{ren2015faster}, region-based fully convolutional networks (R-FCN) \citep{dai2016r}, Mask R-CNN \citep{he2017mask}, Cascade R-CNN \citep{cai2018cascade}, etc. Faster R-CNN is the classic object detector, whose main contribution is the region proposal network (RPN) that is almost costless. 
        In order to further improve the efficiency of Faster R-CNN, R-FCN utilizes a position-sensitive score map and a location-sensitive region of interest (RoI) pooling layer to retain spatial information. 
        As for one-stage detectors, one of the typical methods is "you only look once" (YOLO) \citep{redmon2016you}. As the originator of one-stage object detectors, YOLO has reached actual real-time performance. Recently, YOLOv7 \citep{wang2022yolov7} has surpassed previous object detectors with excellent detection accuracy and real-time performance. 
        In addition to YOLO series methods, one-stage detectors also include single shot multi-box detector (SSD) \citep{liu2016ssd}, RetinaNet \citep{lin2017focal}, EfficientDet \citep{tan2020efficientdet}, etc.
        Due to the proposal of feature pyramid network (FPN) \citep{lin2017feature} and focal loss, scholars have paid attention to anchor-free object detection methods, which are divided into two branches. The first type is the dense prediction-based methods represented by DenseBox \citep{huang2015densebox} and FCOS \citep{tian2019fcos, tian2020fcos}, which densely predict the relative position of anchor frames. Another is the keypoint-based methods represented by CornerNet \citep{law2018cornernet} and CenterNet \citep{zhou2019objects}. The essential difference between anchor-based and anchor-free methods is the mode definition of positive and negative training samples, which has a great impact on the detection performance \citep{zhang2020bridging}.
        
        For object detection tasks, the CNN-based methods have dominated for a long time, but they are only partially end-to-end because of their reliance on hand-designed parts, e.g., anchor frame generation, non-maximal suppression (NMS), and post-processing. The transformer \citep{dosovitskiy2020image}, a network architecture based on encoders and decoders, has received widespread attention in recent years. However, CNN-based methods are so mature that it is difficult to replace them with transformers in a short time \citep{arkin2021survey}, and there is a trend to combine transformers with CNN architectures currently.

    \subsection{Object Detection Methods on Maritime UAVs} \label{3.2}
\begin{figure}[t]
	       \centering
	       \includegraphics[width=0.8 \linewidth]{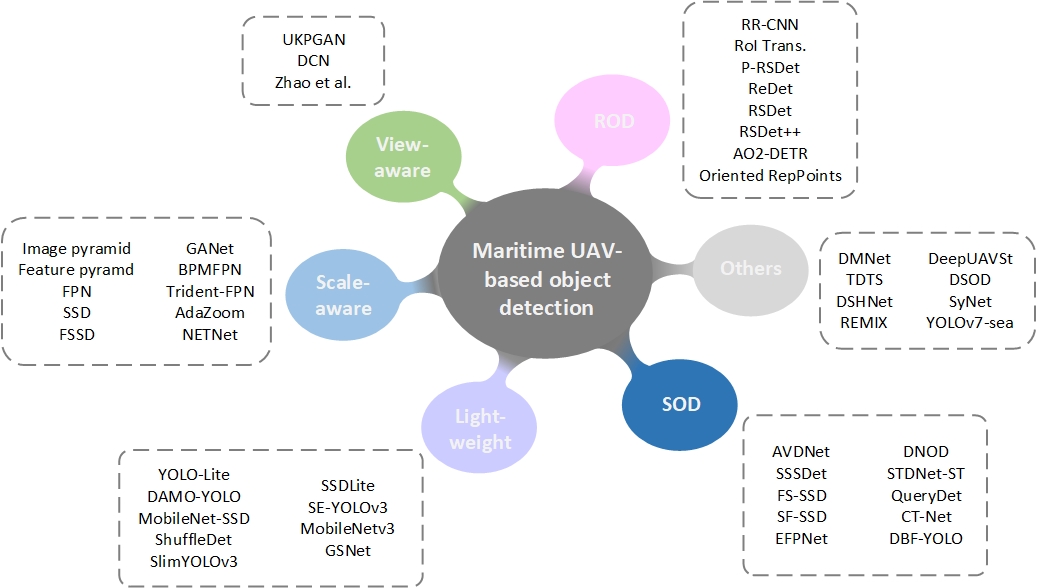}
	       \caption{The structured taxonomy of existing deep learning-based methods for UAV-based object detection, which mainly includes five major methods and others.} 
	       \label{all-methods} 
        \end{figure}

        The generic object detection is not transferable directly on maritime UAV aerial images due to the challenges mentioned in Section \ref{challenge}. Therefore, many methods have been proposed to improve detection accuracy and speed, which are also effective in overcoming these challenges encountered by maritime UAVs. In this section, we review these methods and classify them into six categories, i.e., scale-aware, small object detection (SOD), view-aware, rotated object detection (ROD), lightweight methods, and others. An overview of these methods is illustrated in Fig. \ref{all-methods}.

        \subsubsection{Scale-aware Methods}
            The scale diversity of objects is one of the most challenging problems in object detection tasks, especially for maritime UAVs. The large-scale objects with rich characteristics are easy to be detected. However, it is difficult to detect dense and small-scale objects due to their scant characteristics. The scale-aware methods are used to improve the detection accuracy on objects with different scales. In this part, we first review generic scale-aware methods, and then introduce relative methods to deal with UAV aerial images.

            \begin{figure}[t]
	            \centering
	            \includegraphics[width=0.8 \linewidth]{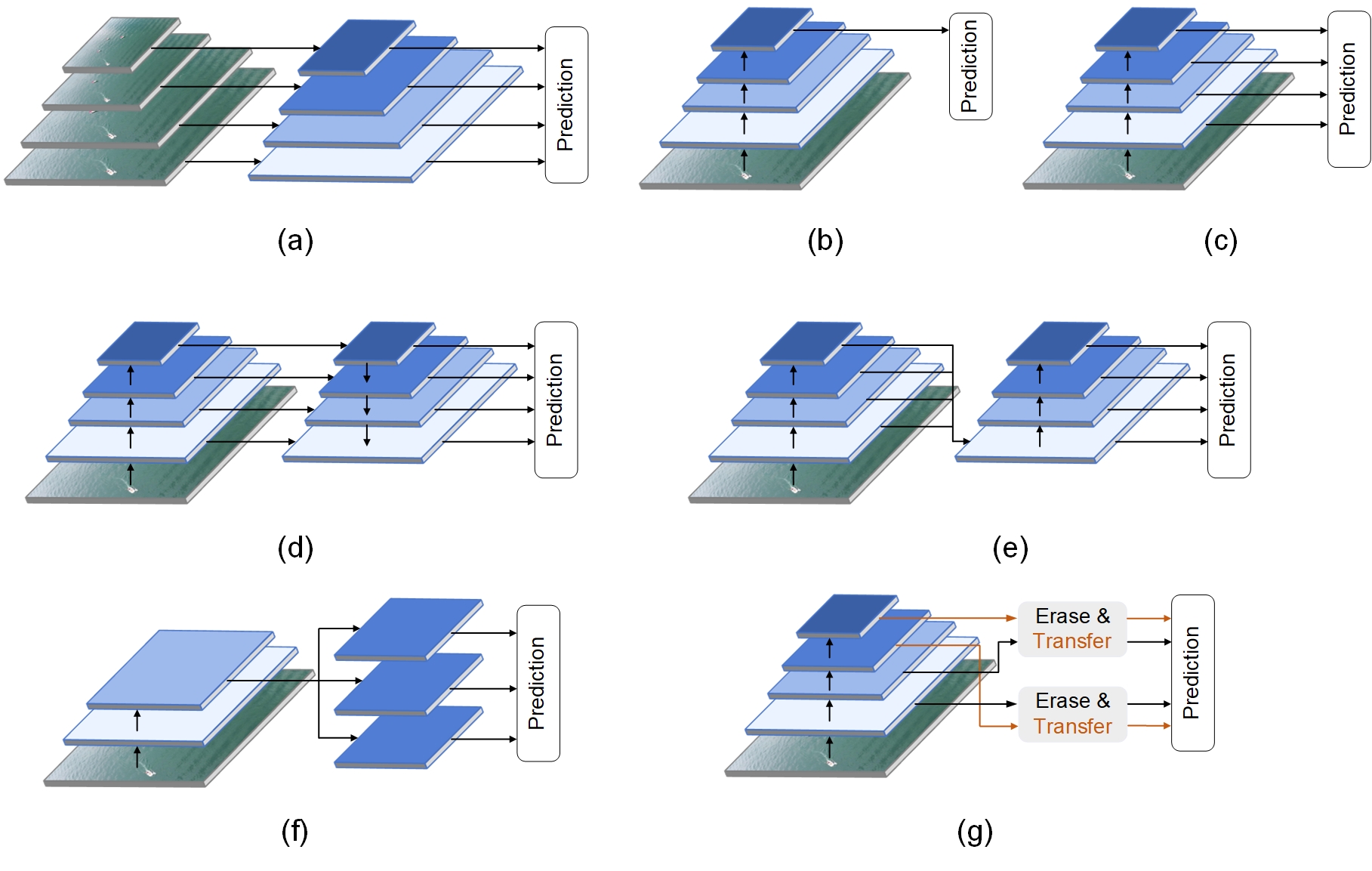}
	            \caption{Some classical network structures for the scale-aware object detection methods. (a) Image pyramid \citep{yin2017multi}. (b) Scale-agnostic method using just one scale feature, such as Faster R-CNN \citep{ren2015faster}. (c) Using the feature pyramid generated from a single CNN, such as the conventional SSD \citep{liu2016ssd}. (d) Feature pyramid, which is the classic scale-aware method, such as the FPN \citep{lin2017feature}. (e) FSSD \citep{li2017fssd}, the improved version of SSD. (f) Trident-FPN \citep{lin2021ecascade}, the multi-branch scale-aware method. (g) NETNet \citep{li2021improving}, used for the feature scale-unmixing.}
	            \label{scale-aware}
            \end{figure}
            The image pyramid \citep{singh2018analysis, yin2017multi} and feature pyramid \citep{lin2017feature} were first proposed to extract features with different object scales. 
            As shown in Fig. \ref{scale-aware} (a), image pyramid methods resize the input image to different scales, which perform competitively within a certain scale range. They utilize the sliding window scheme to detect objects by moving a window of fixed size over the image. However, it requires extremely complex computation, resulting in the poor efficiency.
            To reduce redundant computations and enhance the feature representation, subsequent works tried to build feature maps at different scales, as shown in Fig. \ref{scale-aware} (b). Some classic detectors, e.g., Faster R-CNN, R-FCN, and YOLO, also called scale-agnostic detectors, utilize the final feature map to make predictions. 
            SSD exploits shallower feature maps to detect small objects, and the deeper feature maps for large objects, as depicted in Fig. \ref{scale-aware} (c). 
            The shallow feature maps have tiny receptive fields but abundant information about the position and edges, which are beneficial for the recognition of small objects. Reversely, the deep feature maps contain large receptive fields and extensive semantic information, and they are helpful for recognizing large objects.
            Motivated by the above phenomenon, the feature pyramid network (FPN) introduces a top-down architecture with skip connections to fuse multi-scale features, as shown in Fig. \ref{scale-aware} (d). It significantly improves detection accuracy by taking comprehensive consideration of different scales. Inspired by the FPN, many outstanding methods \citep{li2017fssd, tan2020efficientdet, wang2021progressive} have been proposed in recent years.
            The feature fusion single shot multibox detector (FSSD) \citep{li2017fssd} fuses features from different layers and generates a series of pyramid feature maps, as shown in Fig. \ref{scale-aware} (e). 
            However, some of the feature maps contain redundant information that is useless for feature extraction. In order to reduce the complexity of the detection network and maintain accurate detection, \cite{liang2019small, liu2020small} adopted the deconvolution layer and only used several particularly vital feature maps. 

            In order to tackle the scale variety problem on UAV aerial images, most of scale-aware methods introduce the FPN or its variants \citep{hong2019patch, lin2020novel} into object detectors.        \cite{cai2019guided} proposed a guided attention network (GANet) to fuse different scales of feature maps for rich semantic feature representation. Compared with the method that multiple serial convolutions were stacked to extract features, the bidirectional parallel multi-branch feature pyramid network (BPMFPN) proposed by \cite{fu2021bidirectional} extracted features of different scales by a single path in parallel.
            Trident-FPN \citep{lin2021ecascade} utilized feature maps with different receptive fields and combines them with features of different hierarchies, as shown in Fig. \ref{scale-aware}(f).          \cite{xu2022adazoom} proposed a novel adaptive zoom (AdaZoom) network for large scene object detection. The AdaZoom network actively zoomed the focused regions that were adaptive to scales and object distribution.
            In order to handle the scale variations, \cite{li2021improving} developed a neighbor erasing and transferring (NET) mechanism for feature scale-unmixing. The proposed NET network (NETNet) generated scale-aware features by erasing features of larger objects from shallow layers, rather than fusing profound features into shallow features, as shown in Fig. \ref{scale-aware}(g).

            The scale-aware methods for multi-scale detection have been widely explored now. For the feature extraction of different scales, it does work to simultaneously consider both detailed information in shallow layers and semantic information in deep layers. However, most scale-aware methods usually bring extra computation and hinder real-time detection. Moreover, the combination of features from deep and shallow layers may take features of large objects into shallow layers, which will make a negative effect on small object detection. Therefore, it is an available research trend to find a more efficient fusion strategy.

        \subsubsection{Small Object Detection Methods}
            The detection of small objects has been a standing challenge for UAV-based object detection tasks. Compared with general scenarios, objects in UAV aerial images tend to be considerably smaller. As for small object detection (SOD), there are common problems primarily caused by information loss, noisy feature representation, and a low tolerance for bounding box perturbation. This section will summarize the existing research about SOD in UAV aerial images.

            \begin{figure}[t]
	            \centering
	            \includegraphics[width=0.8 \linewidth]{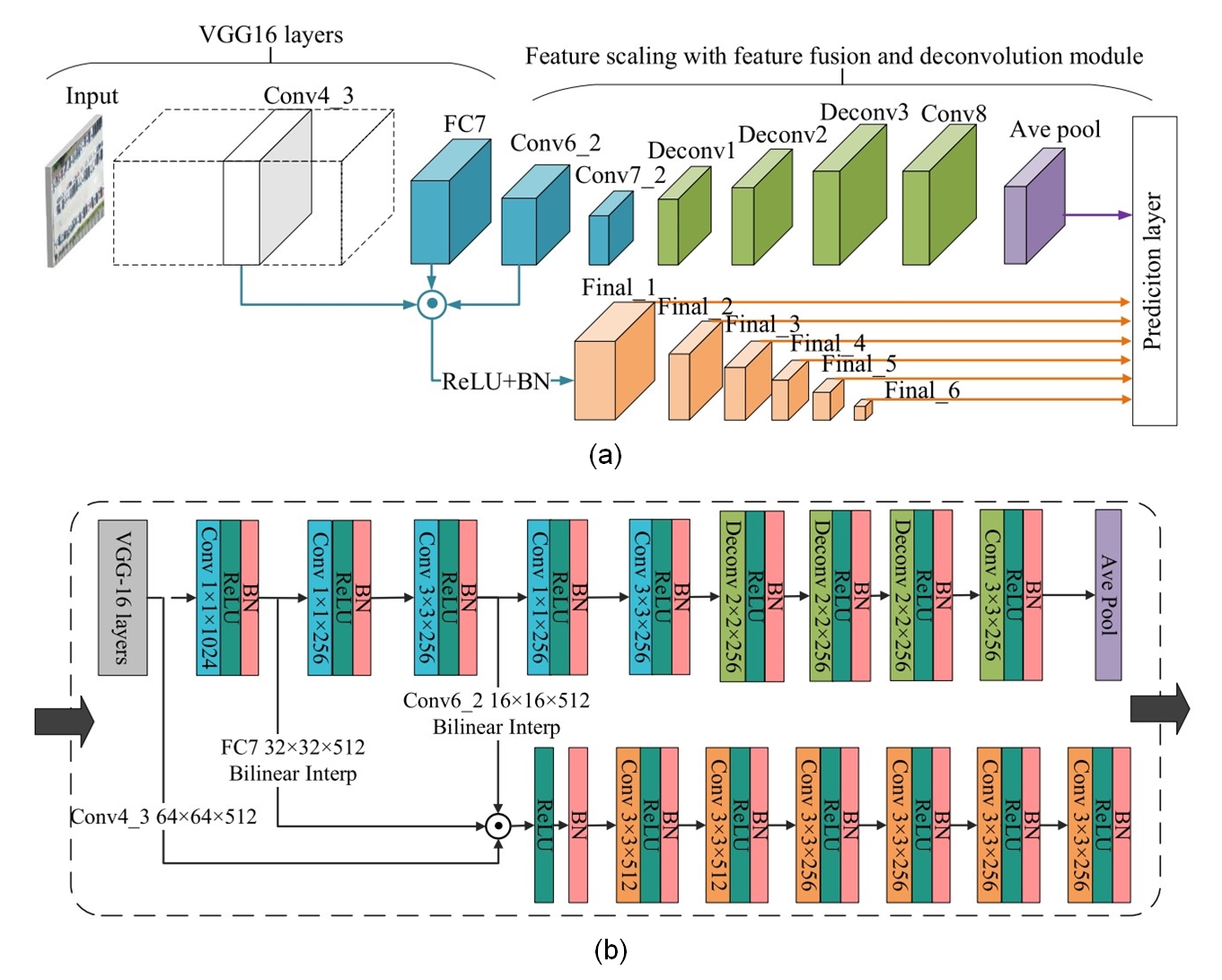}
	            \caption{(a) FS-SSD network \citep{liang2019small} with feature fusion module and average pooling layer. (b) The detailed structure of the additional module.}
	            \label{small_object_methods} 
            \end{figure}
            %

            To deal with the problem of vanishing features about small objects, \cite{mandal2019avdnet} designed an one-stage vehicle detection network (AVDNet), which introduced ConvRes residual blocks at multiple scales. 
            In order to detect small objects under complex backgrounds, \cite{mandal2019sssdet} proposed a simple short and shallow network (SSSDet).  
            A feature fusion and scaling-based single shot detector (FS-SSD) \citep{liang2019small}, as shown in Fig. \ref{small_object_methods}, was proposed for SOD in UAV aerial images. 
            The feature pyramid was formed by adding an extra scaling branch operation, and then the original feature fusion branch was adjusted. To further improve detection accuracy, the spatial context analysis was used to incorporate object spatial relationships into object re-detection.
            Another SSD-based detector (SF-SSD) proposed by \citep{yu2021spatial} leveraged the within-class similarity and provided a new spatial cognition algorithm. 
            The feature fusion idea of the FPN helps to focus on small objects, but the feature coupling at different scales still damages the detection performance. \cite{deng2021extended} proposed an extended feature pyramid network (EFPNet) with an additional high-resolution pyramid level, which was used to super-resolve features and extract credible regional details. 
            A dual neural network object detection (DNOD) was proposed by \cite{tian2021dual}. It was the first sub-network that detected objects with the optimal confidence threshold to ensure high accuracy. The second sub-network screened out objects with a higher confidence threshold to ensure correct detection. The DNOD improved the detection accuracy of small objects while consuming much computing resources. 
            \cite{bosquet2021stdnet} proposed an end-to-end spatio-temporal CNN named STDNet-ST, composing of two branches whose detections were binded by a correlation module to create spatio-temporal small object tubelets.  
            The detection of small objects always leads to expensive calculations. In order to avoid useless calculations of background areas, 
            \cite{yang2022querydet} proposed the QueryDet.
            It used local high-resolution features and then utilized a novel query mechanism to accelerate the inference speed. 
            The convolution-transformer network (CT-Net) \citep{ye2022ct} was proposed to improve the detection performance for multi-scale objects, especially for small objects. Moreover, as one part of CT-Net, a directional feature fusion structure (DFFS) used the element-wise summation operation to fuse the corresponding feature maps. 
            The detection based on shallow feature fusion-YOLO (DBF-YOLO) \citep{liu2023dbf} was an improvement on dealing with the information loss. It retained details of small objects by introducing a shallow feature extraction network.

            In order to extract high quality feature representations for small objects, there are many ways, i.e., super-resolution, attention mechanism, context modeling, and feature fusion strategies in scale-aware methods. However, deepening the backbone network may have a prejudice against high-quality feature extraction for small objects. There is also a huge gap between SOD networks and generic detection models, and many SOD methods inevitably increase the computational burden. In addition, due to the modification of network details, the detection performance of large objects should also be paid attention.

        \subsubsection{View-aware Methods}
            The movement of maritime UAVs and observed objects, as well as the adjustment of camera orientation, lead to multi-view object features. It is one of the unique challenges to detect maritime objects from different aerial views. Nevertheless, the related literature is limited, and most view-aware methods are developed in the face recognition domain. This section will review these view-aware methods that can help to learn multi-view features.

            \begin{figure}[t]
	            \centering
	            \includegraphics[width=0.45 \linewidth]{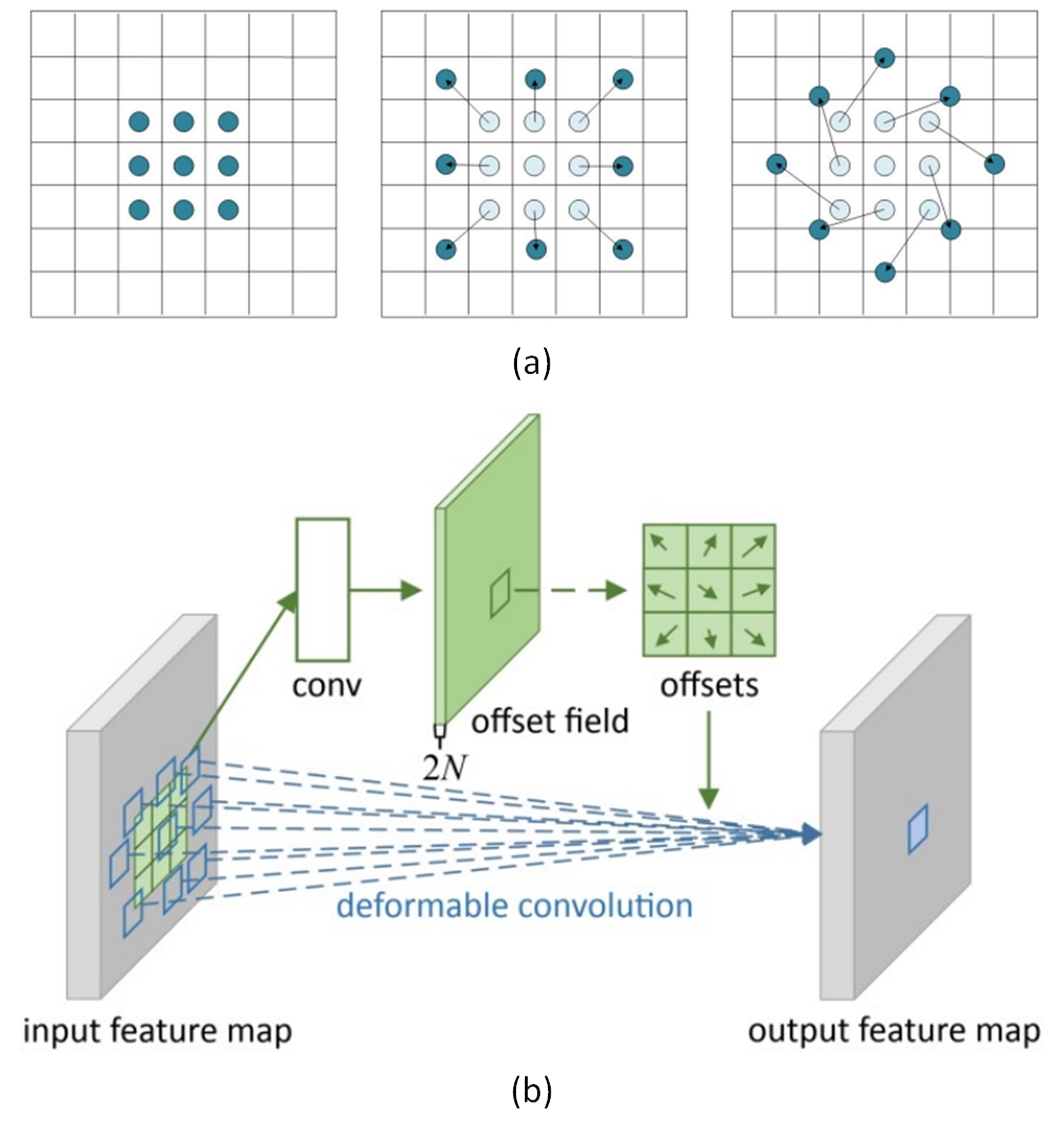}
	            \caption{(a) Deformable characteristics with 3x3 kernel size sample points. From left to right: standard convolution; bigger scale deformable convolution; rotation deformable convolution. (b) The deformable bias layer structure \citep{zhao2020rapid}.}
	            \label{view-aware1}. 
            \end{figure}
            In the maritime UAV aerial images, ships have various attitudes, which is similar to variations of facial postures in the face recognition tasks \citep{du2022elements}. The multi-view face recognition methods will give useful references to the view-aware object detection. There are currently some methods for multi-view feature learning. First, the most direct way to improve the capability of multi-view learning is to obtain enough multi-view data of objects. More data will be generated by generative models, such as the generative adversarial network (GAN). If the training data contains new perspectives and rotation angles of objects, the network can learn much features under various views. However, it is difficult and high-cost to get rich videos and images for some circumstances, i.e., search and rescue missions. Second, it is suitable to use three-dimensional (3D) reconstruction technology when there are a few training samples. Multiple samples of different views will be synthesized by rotating the reconstructed model. It is an effective way to learn multi-view features by utilizing angular invariant features without the effect of postural changes. \cite{you2022ukpgan} proposed a self-supervised 3D key point detector (UKPGAN) where keypoints were detected to reconstruct the original object shape. The detector was rotation invariant without data augmentation, allowing the local keypoint representation to get rid of rotation. The local rotation invariant features were provided through the rotation invariant feature extraction. The CNN shows poor performance on rotation and scale variations, and the deformable convolution is beneficial for modeling arbitrary deformation \citep{ding2019learning}. Under different viewing angles, features extracted by the generic object detection network are non-transferable. In order to improve the transferability of features, the deformable convolutional network (DCN) \citep{dai2017deformable} was used by \cite{zhang2019dense} for feature extraction.
            The deformable convolution block contained an additional convolutional layer to learn the offset, and it improved the adaptability of viewpoint variations. It was hard for the standard convolution to learn variations in rotation views, \cite{zhao2020rapid} got the solution by changing convolution kernel structure which is shown in Fig. \ref{view-aware1} (a). Furthermore, a learnable bias layer was added, offsetting the position of the convolution kernel sample point, as shown in Fig. \ref{view-aware1} (b). In this way, it was more adaptable and achievable for object characteristics and efficient convolution.

        \subsubsection{Rotated Object Detection Methods}
            In UAV aerial images, the rotation of objects varies greatly, which is mainly embodied in the rotated direction. Horizontal bounding boxes contain much background information and cause overlap between adjacent objects, which does damage to the detection accuracy. Objects, e.g., ships and vehicles, contain the orientation information that is always ignored by horizontal bounding boxes. To increase the representation of object localization, rotated detection boxes with the angle information are introduced, especially for the detection in aerial images \citep{yang2021dense, yang2022detecting}.

\begin{figure}[t]
	            \centering
	            \includegraphics[width=1 \linewidth]{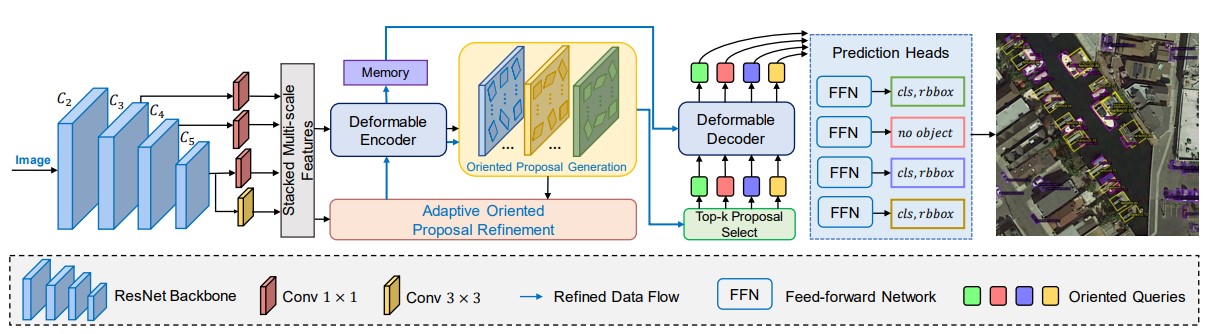}
	            \caption{The illustration of the AO2-DETR method \citep{dai2022ao2}, which adapts the standard Deformable DETR by introducing an oriented proposal generation mechanism, an adaptive oriented proposal refinement module, and a rotation-aware set matching loss.}
	            \label{rotated_object_detection2} 
            \end{figure}

            The research on rotated object detection (ROD) began in the oblique scene text detection \citep{zhou2017east, jiang2017r2cnn}, and then turned to aerial image detection, mainly focusing on the vehicle, UAV, and ship detection. For the ship object detection, \cite{liu2017rotated} proposed a rotated region CNN (RR-CNN), which extracted characteristics of the rotation region and accurately located rotating objects. To learn the transformation parameters, \cite{ding2019learning} proposed a lightweight RoI Transformer (RoI Trans.), applying spatial transformations to the RoI by rotated bounding box annotations. 
            The RoI Trans. can be easily embedded into oriented object detectors. An anchor-free polar remote sensing object detector (P-RSDet) \citep{zhou2020arbitrary} used a simple object representation model. It achieved competitive detection accuracy by prediction of the center point, as well as the regression of one polar radius and two polar angles. 
            \cite{han2021redet} proposed a rotation-equivariant detector namded ReDet, which explicitly encoded the rotation equivariance and invariance. The rotational equivariance network was combined into ReDet to extract rotational equivariance features. It can accurately predict the direction and greatly reduce the model size. Rather than using a five-parameter-based regression with different measurement units, the rotation sensitivity detection network (RSDet) \citep{qian2021learning} reformulated the ROD problem as predicting object corners. 
            Furthermore, a point-based anchorless rotation object detector named RSDet++ \citep{qian2022rsdet++} was proposed to better detect objects smaller than 10 pixels. For better rotation estimation, \cite{dai2022ao2} proposed a transformer-based arbitrary-oriented detector named AO2-DETR, as shown in Fig. \ref{rotated_object_detection2}, eliminating multiple anchors and complex pre/post-processing. The Oriented RepPoints detector proposed by \cite{li2022oriented} detected aerial objects through the adaptive point representation. It was able to capture the geometric information of arbitrary directional instances.

            The recent ROD methods are primarily derived from generic object detectors and introduce the orientation parameter, dominating the research on aerial object detection. However, due to the definitions of the angular orientation and rotated bounding box, the direct orientation prediction encounters the discontinuity of loss and regression inconsistency. From another point of view, rotated object detectors predict aerial images from overhead perspectives, which is also beneficial to improve the accuracy of object detection in UAV aerial images.

        \subsubsection{Lightweight Methods}
            Most deep learning-based object detection methods are executed on GPUs with huge computational power and memory capacity, which are much faster than central processing units (CPUs). However, heavy operations on GPUs always result in large energy consumption and memory storage. The limited computing resources restrict the deployment of large models on maritime UAVs. Many researchers thus began to design lightweight object detection networks based on deep learning technologies.

            \begin{figure}[t]
            	\centering
            	\includegraphics[width=0.8 \textwidth]{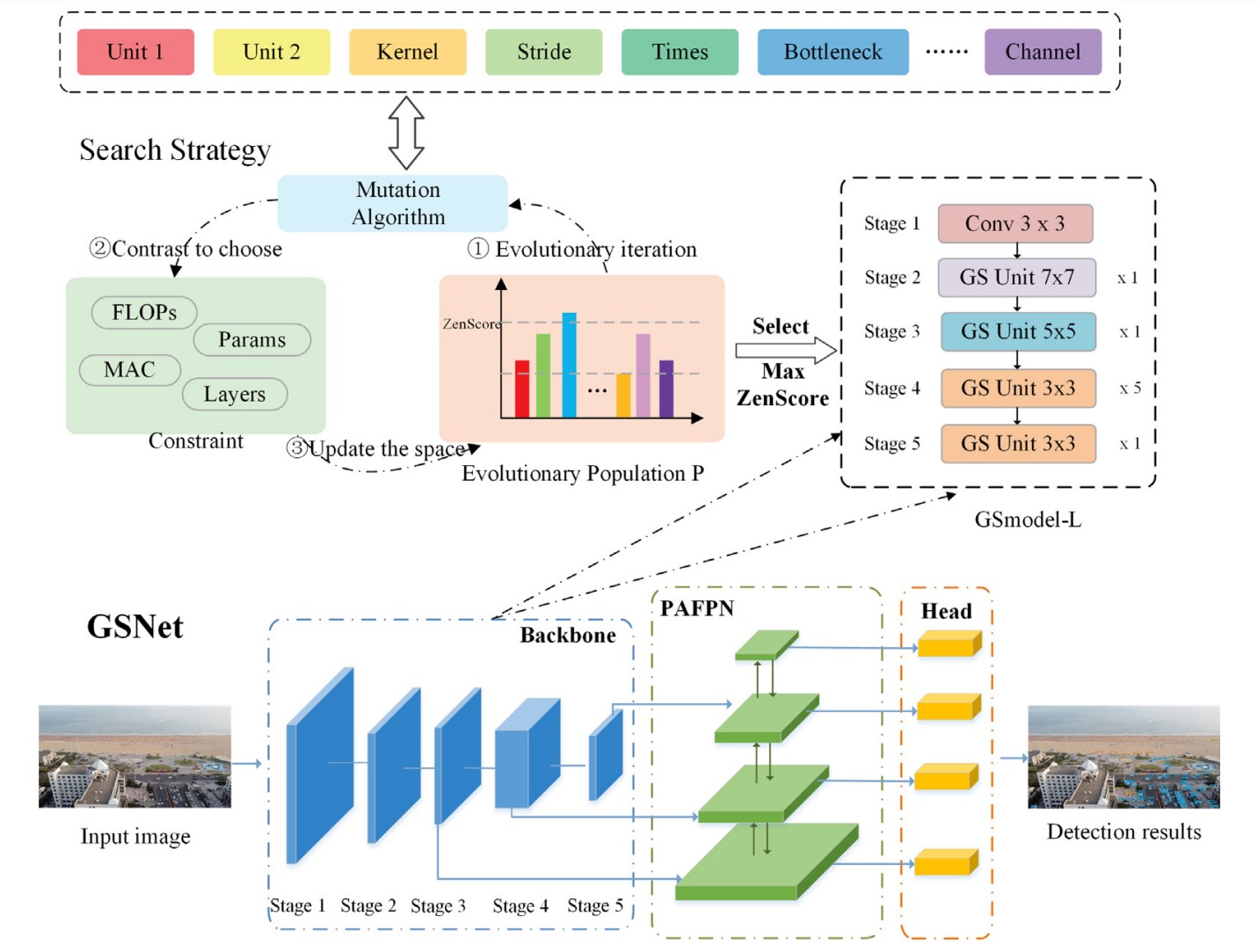}
            	\caption{The pipeline of GSNet \citep{yao2022lightweight}. Firstly, a new lightweight search space is designed. Then, the parameters, FLOPs, layers, and MAC are added to the search strategy as constraints, the optimal network GSmodel-L is searched based on the Zen-NAS evolutionary algorithm. Finally, the best GSmodel-L is used as the backbone and joined with GhostPAN and detection heads to construct the lightweight network.}
            	\label{lightweight} 
            \end{figure}
            There are mainly three different directions to design lightweight neural networks \citep{ge2020survey}, i.e., the manual design of lightweight neural network models, the compression of network models, and the automatic neural network architecture based on the network architecture search (NAS). The manual design of lightweight networks, such as MobileNet \citep{howard2017mobilenets}, ShuffleNet \citep{zhang2018shufflenet}, and SqueezeNet \citep{wu2017squeezedet}, requires more efficient computation modes to perform convolution operations and reduce the number of convolution kernels and channels. Nevertheless, it requires high time costs to design high performance networks. The model compression techniques include network pruning \citep{ye2018rethinking, liu2017learning}, parameter quantization \citep{wu2016quantized}, model distillation \citep{chen2017learning}, etc. Based on these compression methods, some models become lightweight, e.g., YOLO-Lite \citep{huang2018yolo}, DAMO-YOLO \citep{xu2022damo}, and MobileNet-SSD \citep{howard2017mobilenets}. 

            For UAV-based object detection, there are many lightweight methods and most of them are based on SSD and YOLO series methods \citep{razaak2019multi, liu2020issd, feng2022low, liu2023lightweight}. \cite{majid2018shuffledet} proposed the ShuffleDet, which was composed of ShuffleNet and a modified variant of SSD. It adopted channel shuffling and grouped convolution to achieve real-time vehicle detection on UAV aerial images. \cite{zhang2019slimyolov3} presented SlimYOLOv3 through channel pruning of convolutional layers. Compared with YOLOv3, it had fewer trainable parameters and floating point operations. 
            \cite{rabah2020heterogeneous} adopted the SSDLite, also a variant of SSD, which was the ideal detector for computing-resources-limited platforms. The one-stage object detectors are designed to speed up the detection process, but are unfriendly for high-resolution image detection tasks. Therefore, \cite{zhou2021lightweight} proposed the SE-YOLOv3 based on YOLOv3. By constructing an attention-aware feature enhancement module, SE-YOLOv3 enhanced the feature description and suppressed redundancy features with global information. \cite{zhang2021improved} utilized the MobileNetv3 for feature extraction, and achieved better performance in pedestrian detection on UAV aerial images. The networks based on the NAS usually cost enormous computation resources. In order to alleviate the problem, \cite{yao2022lightweight} proposed the ghost shuffle network (GSNet) based on the zero-shot NAS \citep{lin2021zen}, as shown in Fig. \ref{lightweight}. The GSNet contained ghost-shuffle units to reduce parameters and floating point operations. 

            The manually designed deep learning-based object detectors always have inherent redundancy, hindering real-time performance and deployment on UAV platforms. Many researchers adopt the lightweight design philosophy, i.e., model compression, lightweight backbone networks, quantization techniques \citep{rad2021optimized}, etc. These methods significantly improve the detection efficiency but inevitably lose some detection accuracy. More work will focus on the trade-off between detection accuracy and speed, and explore flexible online pruning approaches.

        \subsubsection{Others}
            Besides the object detection methods summarized before, other methods can also achieve competitive detection performance on UAV aerial images. Aiming at the uneven distribution of objects, \cite{li2020density} proposed a density-map guided object detection network (DMNet), utilizing the spatial and context information to improve the detection performance on high-resolution aerial images. \cite{ding2020train} explored the usability of data sparsity and proposed the train in dense and test in sparse (TDTS) method to exploit sparsity. To deal with the problem of the long-tail distribution on UAV aerial images, \cite{yu2021towards} proposed a dual sampler and head detection network (DSHNet). Comprising class-biased samplers and bilateral box heads, the DSHNet coped with tail and head classes in a dual-path manner. \cite{jiang2021flexible} presented a flexible framework named REMIX for high-resolution object detection on UAV platforms. \cite{jain2021ai} proposed the DeepUAVSt to detect objects on UAV aerial images, and adopted an optimization architecture to achieve superior performance. A deeply supervised object detector (DSOD) \citep{shen2017dsod}, as a multi-scale proposal-free detection framework, was entirely trained on UAV aerial images by \cite{micheal2021object}. In order to decrease the high false negative rate of multi-stage detectors and increase the performance of one-stage detectors, an ensemble network named SyNet \citep{albaba2021synet} combined multi-stage and one-stage detectors, and made a prediction by fusing individual predictions of each detector. \cite{zhao2023yolov7} proposed the YOLOv7-sea to detect sea surface tiny objects. The detector searched attention regions in the maritime scene by adding a prediction head and integrating a parameter-free attention module.

\section{UAV-based Datasets}
    The large-scale and well-annotated datasets play a vital role in the development of deep learning-based object detection methods. Profited from UAV aerial photography technology, many UAV-based datasets are built to promote the research of object detection. Ideally, datasets should cover as many scenarios as possible from various view angles, while qualified maritime UAV-based datasets are exceedingly limited. In this section, we provide a review of existing UAV aerial datasets summarized in Table \ref{MS2ship_dataset}. They are categorized into three parts, i.e., general scenarios, maritime scenes, and small objects.

\begin{table}[]
\renewcommand\arraystretch{1.4}
    \begin{center}
        \caption{Commonly used UAV-based datasets for general scenarios, maritime scenes, and small objects.}
\scriptsize
\resizebox{\linewidth}{!}{   
\begin{tabular}{cccccccccc}
\hline
\textbf{Catogory}                                                              & \textbf{Dataset} & \textbf{\begin{tabular}[c]{@{}c@{}}Resolution\\ (pixels)\end{tabular}}                      & \textbf{\begin{tabular}[c]{@{}c@{}}Object \\ Classes\end{tabular}} & \textbf{Images}                                                & \textbf{Videos}                                                              & \textbf{Instances}      & \textbf{\begin{tabular}[c]{@{}c@{}}Altitude\\ (m)\end{tabular}} & \textbf{Year} & \textbf{Public?}                                              \\ \hline
                 & UAV123           & 1280                                                                                        & -                                             & -                                                                & \begin{tabular}[c]{@{}c@{}}123\\ ( \textgreater{}110K\\ frames)\end{tabular} & -                         & 5 to 50                                                         & 2016          & \begin{tabular}[c]{@{}c@{}}
                                                                              {\href{https://cemse.kaust.edu.sa/ivul/uav123}{Yes: Click here}}\end{tabular} \\ \cline{2-10} 
\multirow{5}*{\begin{tabular}[c]{@{}c@{}}General\\ Scenarios\end{tabular}}                                                                               & DTB70            & 1280×720                                                                                    & -                                             & -                                                                 & 70                                                                             & - & 0 to 10                                                         & 2017          & \begin{tabular}[c]{@{}c@{}}{\href{https://github.com/flyers/drone-tracking}{Yes: Click here}}\end{tabular} \\ \cline{2-10} 
                                                                               & DAC-SDC          & 640×360                                                                                     & 12                                                                    & 150K                                                             & -                                                                              & -                         & -                                                               & 2019          & \begin{tabular}[c]{@{}c@{}}{\href{https://github.com/jgoeders/dac_sdc_2021}{Yes: Click here}}\end{tabular} \\ \cline{2-10} 
                                                                               & BirdsEyeView     & \begin{tabular}[c]{@{}c@{}}850×480\\ 1920×1080\end{tabular}                                 & 6                                                                     & 5K                                                               & \textgreater{}70                                                               & 10K                       & -                                                               & 2019          & No                                                            \\ \cline{2-10} 
 & VisDrone2022     & \begin{tabular}[c]{@{}c@{}}Image: \\ 2000×1500 \\ Video: \\ 3840×2160\end{tabular}          & 10                                                                    & 10K                                                              & \begin{tabular}[c]{@{}c@{}}400\\ (\textgreater{}265K \\ frames)\end{tabular}               & \textgreater{}2.6M        & -                                       & 2022          & \begin{tabular}[c]{@{}c@{}}{\href{http://aiskyeye.com/home/}{Yes: Click here}}\end{tabular} \\ \hline
                   & Seagull          & \begin{tabular}[c]{@{}c@{}}1920×1080\\ 1024×768\\ 640×480\\ 384×288\\ 1024×648\end{tabular} & 6                                                                     & -                                                                & \begin{tabular}[c]{@{}c@{}}19\\ (150K \\ frames)\end{tabular}                  & -                         & 150 to 300                                                      & 2017          & \begin{tabular}[c]{@{}c@{}}{\href{https://vislab.isr.tecnico.ulisboa.pt/seagull-dataset/}{Yes: Click here}}\end{tabular} \\ \cline{2-10} 
\multirow{4}{*}{\begin{tabular}[c]{@{}c@{}}Maritime\\ Scenes\end{tabular}}                                                                               & AFO              & \begin{tabular}[c]{@{}c@{}}From \\ 1280×720 \\ to \\ 3840×2160\end{tabular}                 & 6                                                                     & -                                                                & \begin{tabular}[c]{@{}c@{}}50\\ (3647 \\ annotated \\ frames)\end{tabular}     & 40K                       & 30 to 80                                                        & 2021          & \begin{tabular}[c]{@{}c@{}}{\href{http://afo-dataset.pl/download/}{Yes: Click here}}\end{tabular} \\ \cline{2-10} 
                                                                               & MOBDrone         & -                                                                                           & 5                                                                     & -                                                                & \begin{tabular}[c]{@{}c@{}}66\\ (\textgreater{}126K \\ annotated \\ frames)\end{tabular}   & \textgreater{}180K        & 10 to 60                                                        & 2022          & \begin{tabular}[c]{@{}c@{}}{\href{http://aimh.isti.cnr.it/dataset/MOBDrone}{Yes: Click here}}\end{tabular} \\ \cline{2-10} 
    & SeaDronesSee     & \begin{tabular}[c]{@{}c@{}}3840×2160\\ 5456×3632\end{tabular}                               & 6                                                                     & 5630                                                             & 230                                                                            & 400K                      & 5 to 260                                                        & 2022          & \begin{tabular}[c]{@{}c@{}}{\href{https://seadronessee.cs.uni-tuebingen.de.}{Yes: Click here}}\end{tabular} \\ \cline{2-10} 
    & ShipDataset     & \begin{tabular}[c]{@{}c@{}}3840×2160\end{tabular}                               & 1                                                                     & 18K                                                             & 3                                                                            & -                      & 500                                                        & 2023          & \begin{tabular}[c]{@{}c@{}}{\href{https://github.com/NZII/ShipDataset}{Yes: Click here}}\end{tabular} \\ \hline
                                                           \multirow{2}{*}{\begin{tabular}[c]{@{}c@{}}Small \\ Objects\end{tabular}} & TinyPerson                     &  \begin{tabular}[c]{@{}c@{}}From \\ 497×700 \\ to \\ 4064×6354\end{tabular}                  & 2                                                                     & \begin{tabular}[c]{@{}c@{}}2369\\ (1610 \\ labeled)\end{tabular} & -                                                      & \textgreater{}72K         & -                                       & 2019          & \begin{tabular}[c]{@{}c@{}}{\href{https://github.com/ucas-vg/PointTinyBenchmark}{Yes: Click here}}\end{tabular} \\ \cline{2-10} 
     & USC-GRAD-STDdb   & 1280×720                                                                                    & 5                                                                     & -                                                                & 115                                                                            & 56K                       & -                                                               & 2020          & \begin{tabular}[c]{@{}c@{}}{\href{https://gitlab.citius.usc.es/brais.bosquet/USC-GRAD-STDdb}{Yes: Click here}}\end{tabular} \\ \hline
\end{tabular}
}
\end{center}
\label{MS2ship_dataset}
\end{table}

    \subsection{Datasets for UAV Aerial Images/Videos}
        \subsubsection{Datasets for General Scenarios}
            \textbf{UAV123} \citep{mueller2016benchmark} is a high-resolution tracking dataset. It contains 12 types of challenging situations, such as object scale change, aspect ratio change, full occlusion, and so on. The most common challenges are scale variation and aspect ratio change, which account for 89$\%$ and 55$\%$, respectively.

            \textbf{DTB70} \citep{li2017visual} is a highly diverse benchmark video dataset. The visual data are captured in complex and changeable aerial scenes. Under various messy scenes, objects are of different sizes. The main feature is the frequent movement between objects and the UAV. 

            \textbf{DAC-SDC} \citep{xu2019dac} was captured by DJ-Innovations (DJI) UAVs. It contains 150,000 images, and the objects are divided into 12 types and 95 subcategories. The object size of most images is 1-2$\%$ of the overall images, which is one of the dominating characteristics of UAV-based aerial images.

            \textbf{BirdsEyeView} \citep{qi2019birdseyeview} contains 70 videos and 5,000 static images. It is highly diverse with multiple sources, different scenes, view angles, and different heights. The dataset covers many real-life scenarios, such as parking lots, street views, social gatherings, and travel.

            \textbf{VisDrone2022} \citep{zhu2021detection} covers different scenarios, various weather, and lighting conditions. These frames contain objects such as pedestrians, cars, bicycles, and tricycles. The crucial properties, including scene visibility and occlusion condition, are also provided by the annotations.

        \subsubsection{Datasets for Maritime Scenes}
            \begin{figure}[t]
    	       \centering
    	       \includegraphics[width=1 \linewidth]{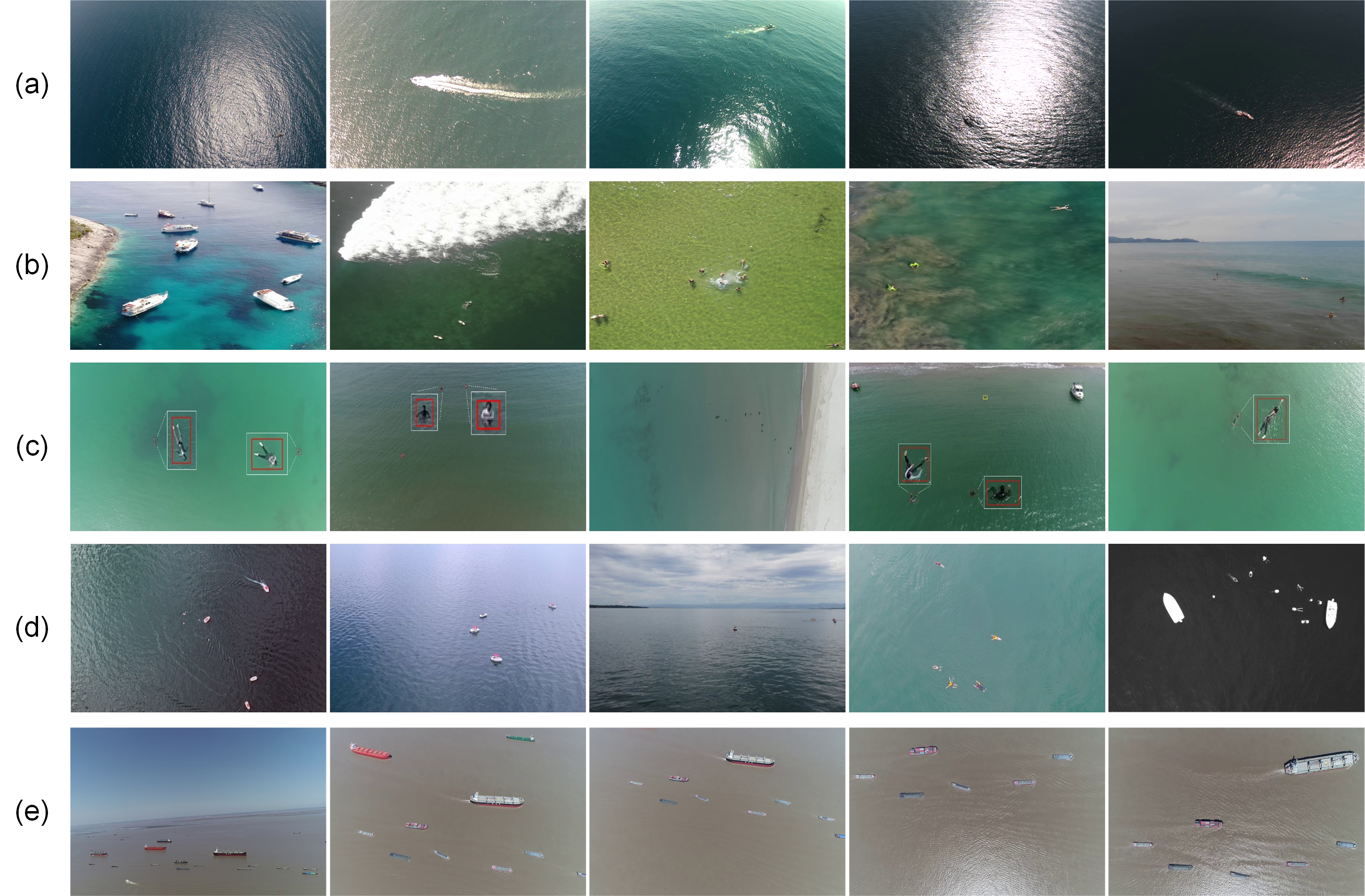}
    	       \caption{Example images from the datasets for maritime scenes. From top to bottom: (a) Seagull dataset \citep{ribeiro2017data}. (b) AFO dataset \citep{gasienica2021ensemble}. (c) MOBDrone dataset \citep{cafarelli2022mobdrone}. (d) SeaDronesSee dataset \citep{varga2022SeaDronesSee}. (e) ShipDataset \citep{zhao2023multi}. Capturing at different altitudes, light conditions, and camera directions, all of these datasets show diverse difficulties, e.g., sun reflections, wakes, multiple boats, scale change, and illumination variations.}
    	       \label{maritime_datasets} 
            \end{figure}
            \textbf{Seagull} \citep{ribeiro2017data} is the first publicly available video dataset applied for maritime surveillance scenarios, as shown in Fig. \ref{maritime_datasets}. It provides seven object types, i.e., cargo ships with 90 meters long, patrol boats with 27 meters long, sailboats, life rafts, hard-shell inflatable boats, oil slicks, and buoys. The dataset supports two tasks that are maritime surveillance/search and environmental monitoring.  

            \textbf{AFO} \citep{gasienica2021ensemble} is proposed for maritime search and rescue applications. A total of 50 videos are captured in different environments and weather conditions, such as water depth/color, coastal shape, and sea conditions, as shown in Fig. \ref{maritime_datasets} (b). The dataset includes 40,000 hand-labeled persons and objects floating on the water, many of which are small and difficult to detect. 

            \textbf{MOBDrone} \citep{cafarelli2022mobdrone} is a video dataset for overboard rescue mission. It contains 66 video clips collected from a UAV flying at 10m-60m. There are five categories, i.e., person, boat, wood, life buoy, and surfboard. Note that 27.72$\%$ of the images have no objects and just contain clear water. The dataset focuses on people who are not wearing life jackets at the sea, and also considers their state of consciousness, as shown in Fig. \ref{maritime_datasets} (c).

            \textbf{SeaDronesSee} \citep{varga2022SeaDronesSee} is proposed for object detection and tracking, as shown in Fig. \ref{maritime_datasets} (d). It provides six categories of labeled objects: swimmers (people without life jackets in the water), floaters (people with life jackets in the water), life jackets, swimmers † (people without life jackets on boats), floaters† (people with life jackets on boats), and small boats.

            \textbf{ShipDataset} \citep{zhao2023multi} covers five scenarios with different shooting angles and lighting conditions in shanghai, China. The ship traffic states in captured images are all congested, which could be seen from Fig. \ref{maritime_datasets} (e). The author used RBoI \citep{chen2017r} as the definition of small object, and this dataset has 96.97$\%$ small ships whose RBoIs range from 0.016$\%$ to 0.51$\%$.

        \subsubsection{Datasets for Small Objects}
\begin{figure}[t]
    	       \centering
    	       \includegraphics[width=1 \linewidth]{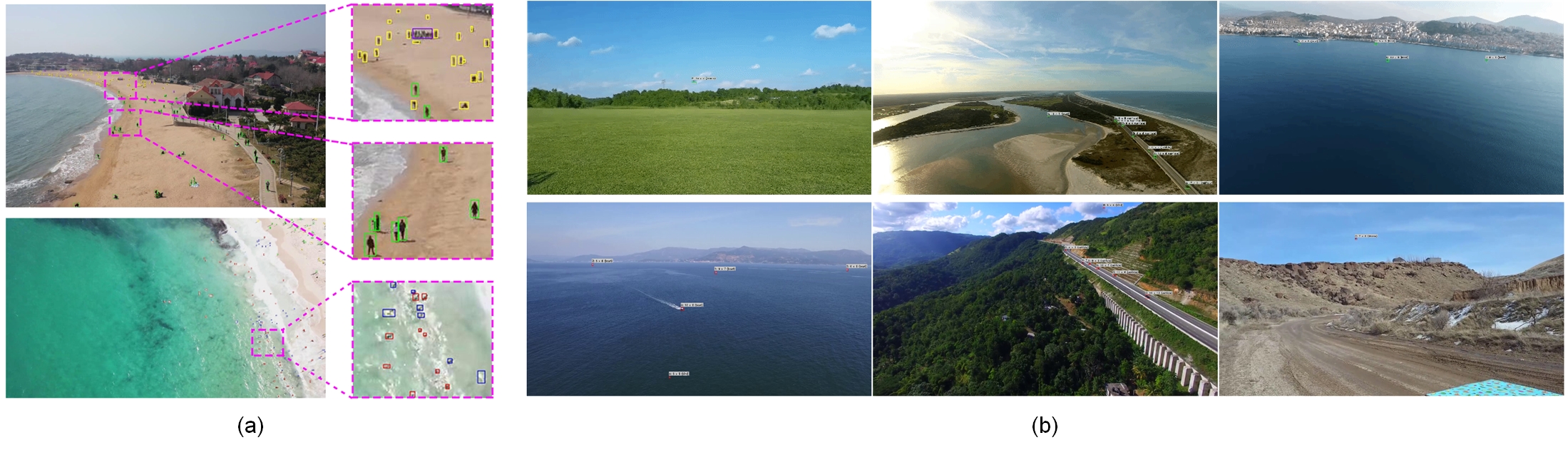}
    	       \caption{Some images from the datasets for small objects. (a) TinyPerson dataset \citep{yu2020scale}. 'earth person', 'sea person', 'uncertain earth person', 'uncertain sea person', and ignore region are represented with green, red, yellow, blue, purple rectangles, respectively. Some regions are zoomed in and shown on the right. (b) USC-GRAD-STDdb dataset \citep{bosquet2020stdnet}. The majority of videos are recorded with a bird eye view over three main landscapes: air, sea, and land.} 
    	       \label{small_object_dataset} 
            \end{figure}

            \textbf{TinyPerson} \citep{yu2020scale} is proposed for the rapid rescue mission and can also be applied for counting people. It contains people in the sea and beach scenes at long distances and focuses on pedestrian detection on the seaside, as shown in Fig. \ref{small_object_dataset} (a). Moreover, the dataset is divided into two subsets, which are small and tiny sets according to size. Due to the extremely tiny size, unidentified areas are assigned with ignore labels. 

            \textbf{USC-GRAD-STDdb} \citep{bosquet2020stdnet} contains 115 video clips and has a sufficient number of small objects with fewer than 16×16 pixels. These videos cover five object categories, i.e., drone, bird, boat, vehicle, and person, as shown in Fig. \ref{small_object_dataset} (b). The pixel areas of these objects range from 16 to 256 pixels.

    \subsection{Our MS2ship Dataset} 
        \begin{figure}[t]
	       \centering
	       \includegraphics[width=1 \linewidth]{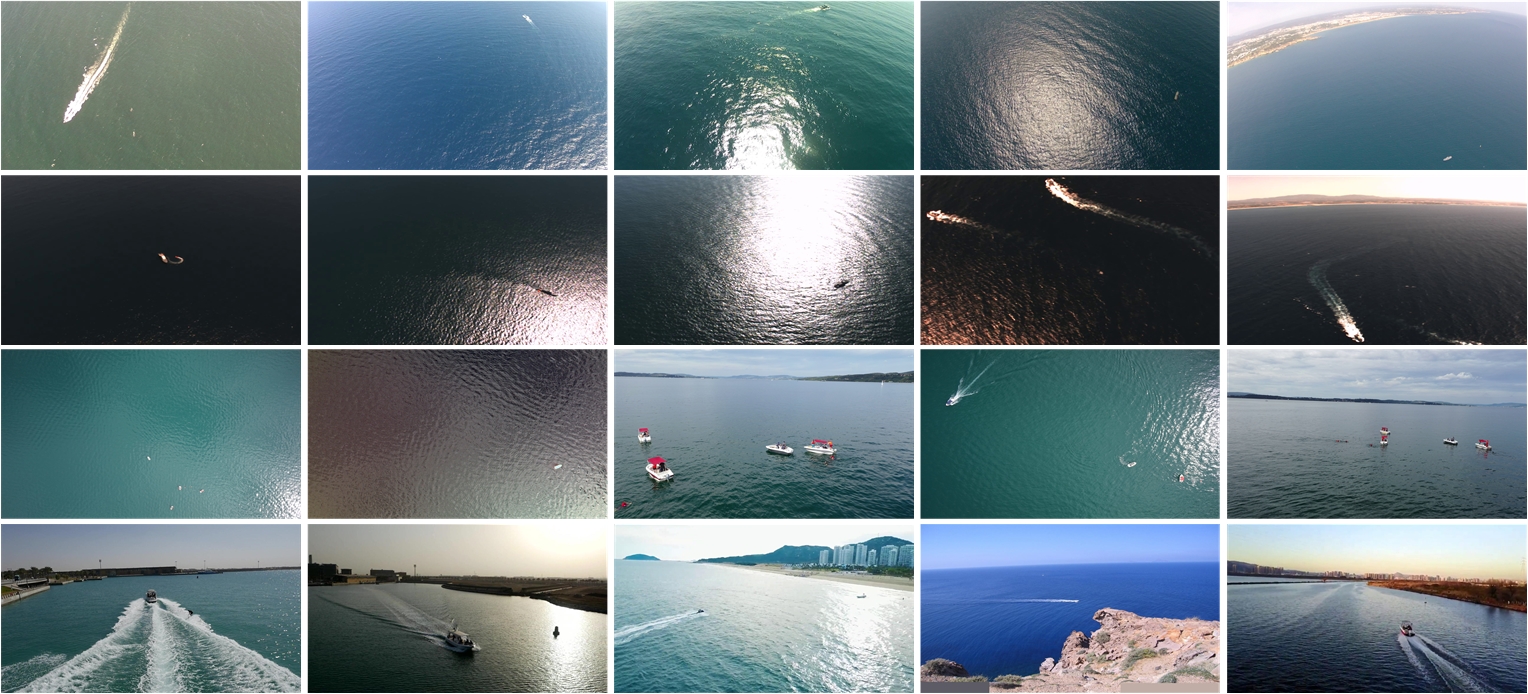}
	       \caption{Some samples of our MS2ship dataset, which contains massive images on the oceans and inland waterways under sunny and low-light conditions.} 
	       \label{MS2ship_figs}
        \end{figure}
        \begin{figure}
	       \centering
	       \includegraphics[width=1 \linewidth]{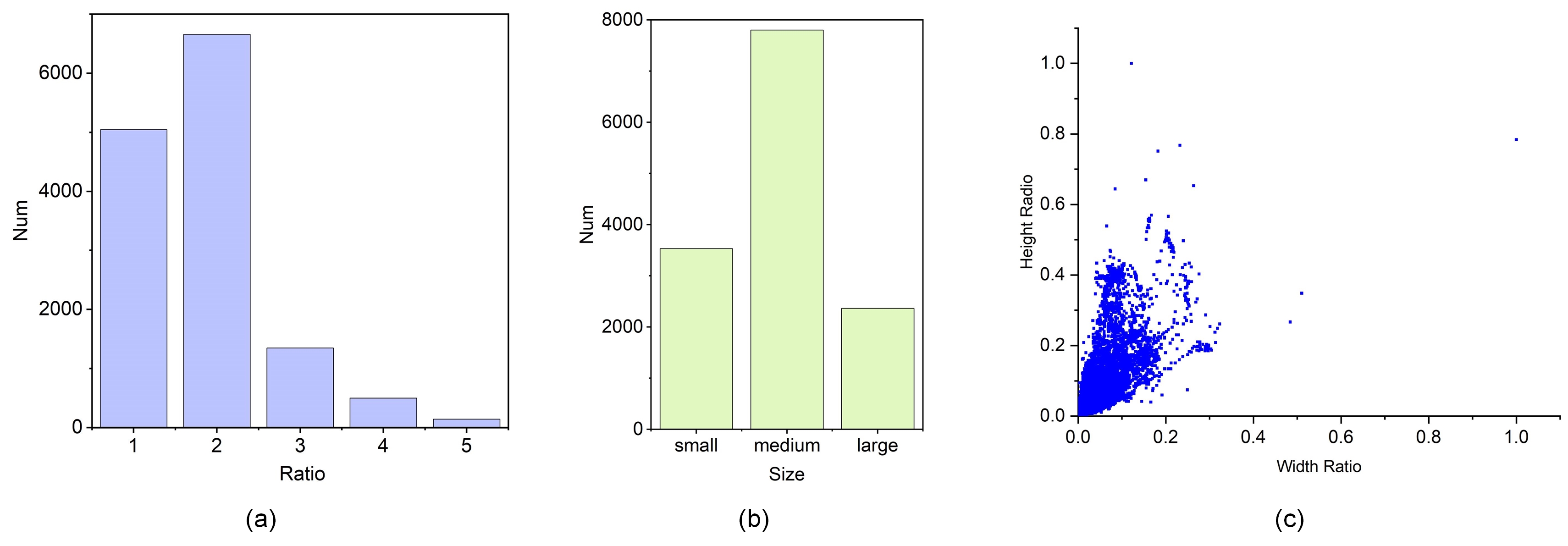}
	       \caption{The analysis of MS2ship. (a) The number of ship objects under different aspect ratios of bounding boxes. (b) The number of ship objects with different sizes, i.e., small, medium, and large size (according to the definition of MS COCO \citep{lin2014microsoft}). (c) The distribution of object sizes. The abscissa and ordinate represent the proportion of the object width and height in the image, respectively.} 
	       \label{dataset_analyse}
        \end{figure}
        In order to compare different maritime object detection methods, it is necessary to create a benchmark dataset for UAV-based ship detection tasks and applications. Therefore, we construct a dataset from the perspective of UAVs. Due to the lack of existing maritime UAV aerial images, we carefully collect publicly available ship images from different sources.
        Our ship images are primarily selected from two maritime datasets that are introduced before, namely, the Seagull and SeaDronesSee datasets. Considering that our maritime ship images are mainly from the two datasets whose names begin with "S", we term the new image dataset as maritime S2 ship (MS2ship). 
        Ideally, the dataset should cover as many scenarios as possible from different capturing views, which is conducive to learning more representative features for the detection model. In order to enrich water scenes of the MS2ship dataset, we carefully picked 425 ship images from UAV123, DTB70, DAC-SDC, and other datasets. The MS2ship dataset ultimately consists of 6,470 images with a total of 13,697 boxes, and there are some representative images in Fig. \ref{MS2ship_figs}. Moreover, the dataset analysis of aspect ratio, sizes, and scale distribution is shown in Fig. \ref{dataset_analyse}. 

        Compared with other datasets, our MS2ship dataset is dedicated to ship detection with three essential characteristics in the maritime scene. Firstly, ship objects are rich in size, shape, and attitude. Secondly, images have unique background features, including back-lighting, poor visibility, and the white wake of ships. Thirdly, our UAV-based single-class object detection dataset is even more challenging, because our task is to detect ships from many complex scenarios and confusing background objects. The MS2ship can be applied to improve the detection accuracy of ship recognition on-board systems. It also assists ship tracking technology to detect the navigation status of ships, suitable for maritime cruising and emergency search and rescue.

\section{Experiment and Discussion}
    In this section, we first introduce the evaluation metrics and experimental implementation details for object detection tasks. Subsequently, a series of experiments on UAV-based datasets are carried out to compare different object detection methods. Finally, we give detailed discussions on the limitations of maritime object detection tasks and future directions for maritime UAVs.  

    \subsection{Evaluation Metrics}
        At present, many typical evaluation metrics are widely applied in object detection tasks. The metrics used in our experiments are presented in this section.

        \textbf{Intersection Over Union:} IoU is defined as the overlap rate between the predicted bounding box and the ground-truth box, that is, the ratio of the intersection and union of the two boxes, as expressed by the Eq. (\ref{IoU}). 
        Where S represents the set of pixels; P represents the prediction result and GT is the ground truth box; |·| is the size of a set; $\cap$ and $\cup$ are the intersection and union operations, respectively.

        \begin{equation}
            IoU=\frac{\left|S_P \cap S_{GT}\right|}{\left|S_P \cup S_{GT}\right|}  ,   
            \label{IoU}
        \end{equation}

        \textbf{Precision and Recall:} Precision denotes the proportion of correct samples in the predicted results, which can be expressed by the Eq. (\ref{precision}). Recall denotes the proportion of positive samples that are correctly detected in the predicted results, as expressed by the Eq. (\ref{recall}). Among the two equations, true positive (TP) means that the true value of the object is true and the model prediction value is true. Similarly, false positive (FP) means that the two values are respectively false and true, and false negative (FN) means that the two values are true and false, respectively. Moreover, the confidence score is the probability that an anchor box contains an object, and the TP and FP are codetermined by the confidence score and IoU. 

        \begin{equation}
            \mathrm{Precision}=\frac{\mathrm{TP}}{\mathrm{TP}+\mathrm{FP}}=\frac{\mathrm{TP}}{\text { All } \text { detections }}  ,
            \label{precision}
        \end{equation}
        \begin{equation}
            \mathrm{Recall}=\frac{\mathrm{TP}}{\mathrm{TP}+\mathrm{FN}}=\frac{\mathrm{TP}}{\text { All } \text { ground truth }}  ,
            \label{recall}
        \end{equation}

        The precision-recall (PR) curve is a plot of precision and recall at varying confidence values, which can also be used to evaluate the performance of a detector. When the confidence score threshold decreases, the recall increases monotonically. However, the precision will fluctuate and tend to decrease overall. 

        \textbf{Average Precision and Mean Average Precision:} AP represents the area under the PR curve, as expressed by Eq. (\ref{AP}), which is the critical metric of object detection methods. To make the equation concise, we use the symbol '\emph{p}' for precision and '\emph{r}' for recall. The mean AP (mAP) is the mean of AP across all object classes, as expressed by Eq. (\ref{mAP}) where '\emph{n}' represents the overall object classes. It is noted that there is only one class in our dataset, i.e., ship, and AP is equal to mAP. According to the COCO benchmark \citep{lin2014microsoft}, the overall AP is the average value over multiple IoU from 0.5 to 0.95 with a step size of 0.05. The $AP_{50}$ is calculated at a single IoU threshold of 0.5. In addition, $AP_S$ represents the average precision at a small scale with sizes less than 32 × 32 pixels.

        \begin{equation}
            A P=\int_0^1 p(r) d r ,
            \label{AP}
        \end{equation}
        \begin{equation}
            m A P=\frac{1}{n}{\sum_{i=1}^n A P_i} , 
            \label{mAP}
        \end{equation}
        \textbf{Average Recall and Mean Average Recall:} AR is the recall averaged over all IoU ranging from 0.5 to 1, which is computed as two times the area under the recall-IoU curve, as expressed by Eq. (\ref{AR}). We use the symbol '\emph{o}' for IoU. The mean AR (mAR) is defined as the mean of AR across all object classes, as expressed by Eq. (\ref{mAR}). In addition, $AR_S$ represents the average recall at a small scale with sizes less than 32 × 32 pixels.

        \begin{equation}
            A R=2 \int_{0.5}^1 \operatorname{recall}(o) d o  ,
            \label{AR}
        \end{equation}
        \begin{equation}
            m A R=\frac{1}{n}{\sum_{i=1}^n A R_i}  .
            \label{mAR}
        \end{equation}
        \textbf{Parameters and GFLOPS:} The evaluation metrics of the model complexity include parameters (Params) and giga floating-point operations per second (GFLOPS). The Params represents the size of the model, while GFLOPS determines its computation complexity. The smaller Params and GFLOPS are, the more conducive to the deployment of object detection models on UAVs.

        \textbf{Frame Per Second:} FPS represents the number of inferred images per second, and expresses how fast the detector is. To make detectors work in real-time, the value of FPS should be increased.

    \subsection{Implementation Details} %
        To conduct fair comparisons of state-of-the-art object detection methods, all experiments on UAV-based dataset are implemented on source code libraries, which are given in the caption of Table \ref{result_1}. We select some classic and advanced methods for comparison, e.g., Faster R-CNN, YOLO series, and CenterNet++ \citep{duan2022centernet++}. 
        Then, We take three benchmark dataset, including the MS2ship, ShipDataset, and AFO dataset, to illustrate the performance of representative object detection methods. Our experiments are carried out on a workbench equipped with an NVIDIA GeForce RTX 2080 Ti, and the operating system is Ubuntu 18.04 LTS. Noting that the detection results are mapped to original images, 
        on which non-maximum suppression (NMS) is performed to prune out redundant predictions and some detection boxes with low confidence are removed. The threshold of confidence score and NMS for all models in our experiments are set to 0.5. Then, other hyper-parameters have the same default values as the source codes. Besides, we use pre-trained backbone models provided by corresponding source codes. In this way, the training speed is increased, as well as the detection accuracy through the transfer learning. All methods use the same augmentation strategy, including mosaic and left-right flip. Moreover, we give the values of $AP_S$ and $AR_S$ to evaluate the performance of SOD.

    \subsection{Experimental Results} 
        To evaluate different object detection models on the UAV-based datasets, we conduct a series of experiments with state-of-the-art methods, which are roughly divided into three categories. The first part contains six generic object detectors, and the second part includes four lightweight detectors. The final part represents two improved detectors used to detect UAV aerial images. This section describes the top-performing methods in each category in detail, and then mediocre methods are stated briefly. 

        \subsubsection{Benchmarking Results}
            Table \ref{result_1}, Table \ref{result_2}, and Table \ref{result_3} report the numerical results of 12 representative methods, respectively on the MS2ship dataset, ShipDataset, and AFO dataset. We list evaluation values on trainval and test datasets, and in the the Table \ref{result_1}, we give other information, e.g., backbone, image size, and Params of inference models. The comparison for detection accuracy, speed, and model size is mainly provided in this part.

\begin{table}[] 
	\renewcommand\arraystretch{1.1}
	\caption{\centering Comparisons of object detection methods on our MS2ship dataset. The \textbf{bold} indicates that the value of metrics ranks in the top 2 of all detectors, and the \underline{italics} indicates that the value ranks at the bottom. In addition, the open source codes are Faster RCNN \citep{ren2015faster}, SSD \citep{liu2016ssd}, RetinaNet \citep{lin2017focal}, YOLOX \citep{ge2021yolox}, CenterNet++ \citep{duan2022centernet++}, YOLOv7 \citep{wang2022yolov7}, YOLOv5 \citep{Jocher_YOLOv5_by_Ultralytics_2020}, PP-PicoDet \citep{yu2021pp}, NanoDet \citep{=nanodet}, TPH-YOLOv5 \citep{zhu2021tph}, and UAV-YOLO \citep{shen2022object}. The methods and their cites in subsequent experiments are the same, which will not be repeated.}
	\footnotesize
	\resizebox{\textwidth}{!}{
		\begin{tabular}{ccccccccccccccccc}
			\hline		
						{\color[HTML]{333333} }                                                                                  & {\color[HTML]{333333} }                                                           & {\color[HTML]{333333} }                                                                                 & \multicolumn{5}{c}{{\color[HTML]{333333} Trainval}}                                                                                                                                                                                                                                                               & \multicolumn{5}{c}{{\color[HTML]{333333} Test}}                                                                                                                                                                                                                             & {\color[HTML]{333333} }                               & {\color[HTML]{333333} }                                                                               & {\color[HTML]{333333} }                                  & {\color[HTML]{333333} }                                \\ \cline{4-13}
						{\color[HTML]{333333} }                                                                                  & {\color[HTML]{333333} }                                                           & {\color[HTML]{333333} }                                                                                 & {\color[HTML]{333333} }                              & {\color[HTML]{333333} }                                & {\color[HTML]{333333} }                               & {\color[HTML]{333333} }                                            & {\color[HTML]{333333} }                               & {\color[HTML]{333333} }                              & {\color[HTML]{333333} }                                & {\color[HTML]{333333} }                               & {\color[HTML]{333333} }                              & {\color[HTML]{333333} }                               & {\color[HTML]{333333} }                               & {\color[HTML]{333333} }                                                                               & {\color[HTML]{333333} }                                  & {\color[HTML]{333333} }                                \\
						\multirow{-3}{*}{{\color[HTML]{333333} Methods}}                                              & \multirow{-3}{*}{{\color[HTML]{333333} Backbone}}                        & \multirow{-3}{*}{{\color[HTML]{333333} \begin{tabular}[c]{@{}c@{}}Input \\ size\end{tabular}}} & \multirow{-2}{*}{{\color[HTML]{333333} AP}} & \multirow{-2}{*}{\cellcolor[HTML]{FFFFFF}{\color[HTML]{333333} $AP_{50}$}} & \multirow{-2}{*}{{\color[HTML]{333333} $AP_S$}} & \multirow{-2}{*}{{\color[HTML]{333333} AR}}               & \multirow{-2}{*}{{\color[HTML]{333333} $AR_S$}} & \multirow{-2}{*}{{\color[HTML]{333333} AP}} & \multirow{-2}{*}{{\color[HTML]{333333} $AP_{50}$}} & \multirow{-2}{*}{{\color[HTML]{333333} $AP_S$}} & \multirow{-2}{*}{{\color[HTML]{333333} AR}} & \multirow{-2}{*}{{\color[HTML]{333333} $AR_S$}} & \multirow{-3}{*}{{\color[HTML]{333333} FPS}} & \multirow{-3}{*}{{\color[HTML]{333333} \begin{tabular}[c]{@{}c@{}}Params\\ /M\end{tabular}}} & \multirow{-3}{*}{{\color[HTML]{333333} GFLOPS}} & \multirow{-3}{*}{{\color[HTML]{333333} Year}} \\ \hline
						\cellcolor[HTML]{FFFFFF}{\color[HTML]{333333} \begin{tabular}[c]{@{}c@{}}Faster\\ RCNN\end{tabular}} & {\color[HTML]{333333} \begin{tabular}[c]{@{}c@{}}Resnet50\\ -FPN\end{tabular}}    & {\color[HTML]{333333} 640}                                                                              & {\color[HTML]{333333} 79.5}                          & {\color[HTML]{333333} \textbf{99.0}}                                             & {\color[HTML]{333333} \textbf{65.6}}                  & {\color[HTML]{333333} 82.3}                                        & {\color[HTML]{333333} \textbf{70.1}}                  & {\color[HTML]{333333} 77.7}                          & {\color[HTML]{333333} 97.8}                            & {\color[HTML]{333333} 63.4}                           & {\color[HTML]{333333} 80.7}                          & {\color[HTML]{333333} 68.2}                           & {\color[HTML]{333333} 33.88}                          & {\color[HTML]{333333} 41.53}                                                                          & {\color[HTML]{333333} 91.41}                             & {\color[HTML]{333333} 2016}                            \\
						\cellcolor[HTML]{FFFFFF}{\color[HTML]{333333} SSD}                                                       & {\color[HTML]{333333} \begin{tabular}[c]{@{}c@{}}Resnet50\\ -FPN\end{tabular}}    & {\color[HTML]{333333} 512}                                                                              & {\color[HTML]{333333} 74.9}                          & {\color[HTML]{333333} 97.6}                                                    & {\color[HTML]{333333} \underline{56.2}}            & {\color[HTML]{333333} 79.4}                                        & {\color[HTML]{333333} 64.5}                           & {\color[HTML]{333333} 73.0}                            & {\color[HTML]{333333} 96.6}                            & {\color[HTML]{333333} \underline{55.0}}              & {\color[HTML]{333333} \underline{77.3}}           & {\color[HTML]{333333} \underline{62.8}}            & {\color[HTML]{333333} 56.31}                          & {\color[HTML]{333333} 24.39}                                                                          & {\color[HTML]{333333} 87.72}                             & {\color[HTML]{333333} 2016}                            \\
						\cellcolor[HTML]{FFFFFF}{\color[HTML]{333333} RetinaNet}                                                 & {\color[HTML]{333333} \begin{tabular}[c]{@{}c@{}}Resnet50\\ -FPN\end{tabular}}    & {\color[HTML]{333333} 640}                                                                              & {\color[HTML]{333333} 78.4}                          & {\color[HTML]{333333} 98.3}                                                    & {\color[HTML]{333333} 58.0}                             & {\color[HTML]{333333} 82.9}                                        & {\color[HTML]{333333} 69.2}                           & \cellcolor[HTML]{FFFFFF}{\color[HTML]{333333} 76.3}  & {\color[HTML]{333333} \textbf{98.2}}                   & {\color[HTML]{333333} 59.1}                           & {\color[HTML]{333333} 81.4}                          & {\color[HTML]{333333} \textbf{68.7}}                  & {\color[HTML]{333333} 23.74}                          & {\color[HTML]{333333} 37.74}                                                                          & {\color[HTML]{333333} 95.66}                             & {\color[HTML]{333333} 2018}                            \\
						\cellcolor[HTML]{FFFFFF}{\color[HTML]{333333} YOLOX-L}                                                   & {\color[HTML]{333333} \begin{tabular}[c]{@{}c@{}}Modified \\ CSP v5\end{tabular}} & {\color[HTML]{333333} 640}                                                                              & {\color[HTML]{333333} 78.7}                          & {\color[HTML]{333333} 98.0}                                                      & {\color[HTML]{333333} 64.9}                           & {\color[HTML]{333333} 81.4}                                        & {\color[HTML]{333333} 68.9}                           & {\color[HTML]{333333} 77.4}                          & {\color[HTML]{333333} 97.9}                            & {\color[HTML]{333333} \textbf{64.1}}                  & {\color[HTML]{333333} 80.2}                          & {\color[HTML]{333333} 68.5}                           & {\color[HTML]{333333} 4.65}                           & {\color[HTML]{333333}  \underline{54.15}}                                                           & {\color[HTML]{333333} \underline{155.31}}             & {\color[HTML]{333333} 2021}                            \\
						\cellcolor[HTML]{FFFFFF}{\color[HTML]{333333} CenterNet++}                                               & {\color[HTML]{333333} Resnet50}                                                   & {\color[HTML]{333333} 512}                                                                              & {\color[HTML]{333333} 77.3}                          & {\color[HTML]{333333} 98.0}                                                      & {\color[HTML]{333333} \textbf{65.9}}                  & {\color[HTML]{333333} 81.9}                                        & {\color[HTML]{333333} \textbf{69.5}}                  & {\color[HTML]{333333} 75.6}                          & {\color[HTML]{333333} 97.8}                            & {\color[HTML]{333333} \textbf{65.4}}                  & {\color[HTML]{333333} 80.3}                          & {\color[HTML]{333333} \textbf{69.4}}                  & {\color[HTML]{333333} 8.23}                           & {\color[HTML]{333333} 45.18}                                                                          & {\color[HTML]{333333} 49.83}                             & {\color[HTML]{333333} 2022}                            \\
						\cellcolor[HTML]{FFFFFF}{\color[HTML]{333333} YOLOv7}                                                    & {\color[HTML]{333333} \begin{tabular}[c]{@{}c@{}}ELANNet-\\ PaFPN\end{tabular}}   & {\color[HTML]{333333} 640}                                                                              & {\color[HTML]{333333} \textbf{82.2}}                 & {\color[HTML]{333333} \textbf{99.5}}                                           & {\color[HTML]{333333} -}                              & {\color[HTML]{333333} \textbf{99.0}}                                 & {\color[HTML]{333333} -}                              & {\color[HTML]{333333} 80.6}                          & {\color[HTML]{333333} \textbf{98.0}}                     & {\color[HTML]{333333} -}                              & {\color[HTML]{333333} \textbf{97.2}}                 & {\color[HTML]{333333} -}                              & {\color[HTML]{333333} 80.87}                          & {\color[HTML]{333333} 36.48}                                                                          & {\color[HTML]{333333} 103.30}                             & {\color[HTML]{333333} 2022}                            \\
						\cellcolor[HTML]{FFFFFF}{\color[HTML]{333333} YOLOv5n}                                                   & {\color[HTML]{333333} \begin{tabular}[c]{@{}c@{}}CSP-\\ Darknet53\end{tabular}}   & {\color[HTML]{333333} 640}                                                                              & {\color[HTML]{333333} \textbf{82.8}}                 & {\color[HTML]{333333} 97.6}                                                    & {\color[HTML]{333333} -}                              & {\color[HTML]{333333} 95.3}                                        & {\color[HTML]{333333} -}                              & {\color[HTML]{333333} \textbf{81.7}}                 & {\color[HTML]{333333} 97.5}                            & {\color[HTML]{333333} -}                              & {\color[HTML]{333333} \textbf{95.2}}                 & {\color[HTML]{333333} -}                              & {\color[HTML]{333333} \textbf{222.22}}                & {\color[HTML]{333333} \textbf{1.76}}                                                                  & {\color[HTML]{333333} \textbf{4.20}}                      & {\color[HTML]{333333} 2020}                            \\
						\cellcolor[HTML]{FFFFFF}{\color[HTML]{333333} \begin{tabular}[c]{@{}c@{}}YOLOX-\\ tiny\end{tabular}}     & {\color[HTML]{333333} \begin{tabular}[c]{@{}c@{}}Modified \\ CSP v5\end{tabular}} & {\color[HTML]{333333} 416}                                                                              & {\color[HTML]{333333} 75.7}                          & {\color[HTML]{333333} 97.0}                                                      & {\color[HTML]{333333} 59.4}                           & \cellcolor[HTML]{FFFFFF}{\color[HTML]{333333} \underline{78.3}} & {\color[HTML]{333333} \underline{63.4}}            & \cellcolor[HTML]{FFFFFF}{\color[HTML]{333333} 74.8}  & {\color[HTML]{333333} 97.9}                            & {\color[HTML]{333333} 59.5}                           & {\color[HTML]{333333} \underline{77.3}}           & {\color[HTML]{333333} 64.0}                             & {\color[HTML]{333333} 31.27}                          & {\color[HTML]{333333} 5.03}                                                                           & {\color[HTML]{333333} 6.40}                               & {\color[HTML]{333333} 2021}                            \\
						\cellcolor[HTML]{FFFFFF}{\color[HTML]{333333} \begin{tabular}[c]{@{}c@{}}PP-PicoDet\\ -L\end{tabular}}   & {\color[HTML]{333333} ESNet}                                                      & {\color[HTML]{333333} 640}                                                                              & {\color[HTML]{333333} 78.4}                          & {\color[HTML]{333333} 97.7}                                                    & {\color[HTML]{333333} 60.9}                           & {\color[HTML]{333333} 81.2}                                        & {\color[HTML]{333333} 65.7}                           & {\color[HTML]{333333} 77.0}                            & {\color[HTML]{333333} 97.5}                            & {\color[HTML]{333333} 60.2}                           & {\color[HTML]{333333} 80.3}                          & {\color[HTML]{333333} 66.1}                           & {\color[HTML]{333333} 31.45}                          & {\color[HTML]{333333} 3.30}                                                                            & {\color[HTML]{333333} 8.30}                               & {\color[HTML]{333333} 2021}                            \\
						\cellcolor[HTML]{FFFFFF}{\color[HTML]{333333} \begin{tabular}[c]{@{}c@{}}NanoDet\\ -Plus\end{tabular}}   & {\color[HTML]{333333} \begin{tabular}[c]{@{}c@{}}Shuffle-\\ NetV2\end{tabular}}   & {\color[HTML]{333333} 416}                                                                              & {\color[HTML]{333333} \underline{64.3}}           & {\color[HTML]{333333} \underline{89.2}}                                     & {\color[HTML]{333333} \underline{36.7}}            & {\color[HTML]{333333} \underline{68.2}}                         & {\color[HTML]{333333} \underline{41.8}}            & {\color[HTML]{333333} \underline{62.6}}           & {\color[HTML]{333333} \underline{89.8}}             & {\color[HTML]{333333} \underline{34.8}}            & {\color[HTML]{333333} \underline{66.6}}           & {\color[HTML]{333333} \underline{40.2}}            & {\color[HTML]{333333} \textbf{88.07}}                 & {\color[HTML]{333333} \textbf{2.44}}                                                                  & {\color[HTML]{333333} \textbf{2.97}}                     & {\color[HTML]{333333} 2021}                            \\
						\cellcolor[HTML]{FFFFFF}{\color[HTML]{333333} \begin{tabular}[c]{@{}c@{}}TPH-\\ YOLOv5\end{tabular}}     & {\color[HTML]{333333} \begin{tabular}[c]{@{}c@{}}CSP-\\ Darknet53\end{tabular}}   & {\color[HTML]{333333} 640}                                                                              & {\color[HTML]{333333} \textbf{82.8}}                 & {\color[HTML]{333333} 98.3}                                                    & {\color[HTML]{333333} -}                              & {\color[HTML]{333333} \textbf{95.9}}                               & {\color[HTML]{333333} -}                              & {\color[HTML]{333333} \textbf{81.4}}                 & {\color[HTML]{333333} 96.2}                            & {\color[HTML]{333333} -}                              & {\color[HTML]{333333} 92.9}                          & {\color[HTML]{333333} -}                              & {\color[HTML]{333333} 44.05}                          & {\color[HTML]{333333} 45.36}                                                                          & {\color[HTML]{333333} \underline{259.60}}              & {\color[HTML]{333333} 2021}                            \\
						{\color[HTML]{333333} UAV-YOLO}                                                                          & {\color[HTML]{333333} \begin{tabular}[c]{@{}c@{}}CSP-\\ Darknet53\end{tabular}}   & {\color[HTML]{333333} 1280}                                                                             & {\color[HTML]{333333} \underline{73.4}}           & {\color[HTML]{333333} \underline{95.2}}                                     & {\color[HTML]{333333} -}                              & {\color[HTML]{333333} 94.7}                                        & {\color[HTML]{333333} -}                              & {\color[HTML]{333333} \underline{71.7}}           & {\color[HTML]{333333} \underline{95.0}}               & {\color[HTML]{333333} -}                              & {\color[HTML]{333333} 93.9}                          & {\color[HTML]{333333} -}                              & {\color[HTML]{333333} 70.80}                           & {\color[HTML]{333333}  \underline{47.36}}                                                           & {\color[HTML]{333333} 115.80}                             & {\color[HTML]{333333} 2022}                            \\ \hline
		\end{tabular}
	}
	\label{result_1}
\end{table}
            \begin{table}[] 
	\renewcommand\arraystretch{1.1}
	\caption{\centering Comparisons of object detection methods on the ShipDataset. The \textbf{bold} indicates that the value of metrics ranks in the top 2 of all detectors, and the \underline{italics} indicates that the value ranks at the bottom.}
	\footnotesize
		\begin{tabular}{ccccccccccc}
\hline
                                   & \multicolumn{5}{c}{Trainval}                                                                                                                                                                                                  & \multicolumn{5}{c}{Test}                                                                                                                                                                                                      \\ \cline{2-11} 
                                   &                               &                                 &                                &                               &                                &                               &                                 &                                &                               &                                \\
\multirow{-3}{*}{Methods} & \multirow{-2}{*}{AP} & \multirow{-2}{*}{$AP_{50}$} & \multirow{-2}{*}{$AP_S$} & \multirow{-2}{*}{AR} & \multirow{-2}{*}{$AR_S$} & \multirow{-2}{*}{AP} & \multirow{-2}{*}{$AP_{50}$} & \multirow{-2}{*}{$AP_S$} & \multirow{-2}{*}{AR} & \multirow{-2}{*}{$AR_S$} \\ \hline
Faster   RCNN-FPN                  & 80.6                                                 & 99.0                                                   & 54.0                                                  & 83.5                          & 62.9                           & 80.0                                                 & 98.1                                                   & 54.2                                                  & 82.4                          & 62.2                           \\
SSD                                & 86.7                                                 & 98.9                                                   & 63.5                                                  & 89.8                          & 70.7                           & 85.3                                                 & 98.9                                                   & 63.2                                                  & 88.7                          & 70.2                           \\
RetinaNet                          & 76.4                                                 & 98.6                                                   & 37.6                                                  & 80.1                          & 53.7                           & 76.0                                                 & 98.6                                                   & 40.4                                                  & 79.5                          & 54.7                           \\
YOLOX-L                            & \underline{71.4}                                           & 96.6                                                   & \underline{27.1}                                            & \underline{73.9}                    & \underline{37.9}                     & \underline{68.5}                                           & \underline{93.5}                                             & \underline{27.6}                                            & \underline{72.6}                    & \underline{37.3}                     \\
CenterNet++                        & 89.6                                                 & 99.0                                                   & \textbf{70.3}                                         & 92.1                          & \textbf{75.5}                  & 88.8                                                 & 99.0                                                   & \textbf{70.8}                                         & 91.2                          & \textbf{75.1}                  \\
YOLOv7                             & 89.9                                                 & \textbf{99.7}                                          & -                                                     & \textbf{99.4}                 & -                              & 88.7                                                 & \textbf{99.6}                                          & -                                                     & \textbf{99.2}                 & -                              \\
YOLOv5n                            & \textbf{91.5}                                        & 99.5                                                   & -                                                     & \textbf{99.3}                 & -                              & \textbf{90.6}                                        & \textbf{99.5}                                          & -                                                     & \textbf{99.1}                 & -                              \\
YOLOX-Tiny                         & \underline{71.2}                                           & \underline{96.5}                                             & \underline{28.1}                                            & \underline{73.9}                    & \underline{37.5}                     & \underline{71.1}                                           & \underline{96.4}                                             & \underline{30.1}                                            & \underline{73.9}                    & \underline{37.3}                     \\
PP-PicoDet-L                       & \textbf{90.7}                                        & 98.9                                                   & \textbf{66.8}                                         & 92.3                          & \textbf{72.9}                  & \textbf{89.7}                                        & 98.9                                                   & \textbf{67.3}                                         & 91.4                          & \textbf{73.0}                  \\
NanoDet-Plus                       & 84.2                                                 & \underline{96.4}                                             & 38.9                                                  & 86.0                          & 44.4                           & 83.1                                                 & \underline{96.4}                                             & 41.8                                                  & 84.9                          & \underline{46.4}                     \\
TPH-YOLOv5                         & 90.3                                                 & 99.3                                                   & -                                                     & 98.8                          & -                              & 89.6                                                 & 99.2                                                   & -                                                     & 98.4                          & -                              \\
UAV-YOLO                           & 86.0                                                 & \textbf{99.6}                                          & -                                                     & 98.7                          & -                              & 86.3                                                 & \textbf{99.5}                                          & -                                                     & 99.0                          & -                              \\ \hline
\end{tabular}
\label{result_2}
\end{table}
\begin{table}[]
\renewcommand\arraystretch{1.2}
\caption{\centering Comparisons of object detection methods on the AFO dataset. The \textbf{bold} indicates that the value of metrics ranks in the top 2 of all detectors, and the \underline{italics} indicates that the value ranks at the bottom.}
\footnotesize
\resizebox{\textwidth}{!}{
\begin{tabular}{ccccccccccccccccc}
\hline
                                                        & \multicolumn{5}{c}{Trainval}                                                                                                                                                               & \multicolumn{11}{c}{Test}                                                                                                                                                                                                                                                                                                                                                                           \\ \cline{2-17} 
                                                        &                     &                        &                       &                      &                       &                       &                       &                      &                      &                       & \multicolumn{6}{c}{AP}                                                                                                                                                                                 \\ \cline{12-17} 
\multirow{-3}{*}{Methods}                               & \multirow{-2}{*}{mAP} & \multirow{-2}{*}{$AP_{50}$} & \multirow{-2}{*}{$AP_S$} & \multirow{-2}{*}{AR} & \multirow{-2}{*}{$AR_S$} & \multirow{-2}{*}{mAP} & \multirow{-2}{*}{$AP_{50}$} & \multirow{-2}{*}{$AP_S$} & \multirow{-2}{*}{AR} & \multirow{-2}{*}{$AR_S$} & \multicolumn{1}{l}{human} & \begin{tabular}[c]{@{}c@{}}wind/\\ sup-board\end{tabular} & \multicolumn{1}{l}{boat} & \multicolumn{1}{l}{bouy} & \multicolumn{1}{l}{sailboat} & \multicolumn{1}{l}{kayak} \\ \hline
\begin{tabular}[c]{@{}c@{}}Faster\\ RCNN\end{tabular}   & 29.6                                         & 63.9                                          & 20.5                                         & \underline{42.4}           & 19.7                  & 29.9                                         & 63.0                                          & \textbf{19.6}                                & 42.1                 & 19.6                  & \underline{19.7}                & \underline{47.2}                                                & \underline{16.2}               & 19.7                     & \underline{17.9}                   & 56.6                      \\
SSD                                                     & 53.5                                         & 91.5                                          & 17.9                                         & 61.2                 & 22.8                  & 52.1                                         & 88.8                                          & 16.7                                         & 60.5                 & \textbf{23.5}         & 38.0                      & 64.0                                                      & 56.5                     & 24.2                     & 63.6                         & 66.2                      \\
RetinaNet                                               & \underline{19.8}                                   & \underline{45.3}                                    & \underline{1.2}                                    & \underline{34.0}           & \underline{2.7}             & \underline{18.2}                                   & \underline{42.3}                                    & \underline{0.9}                                    & \underline{34.6}           & \underline{3.5}             & \underline{15.5}                & \underline{40.0}                                                & \underline{4.5}                & \underline{3.1}                & \underline{0.9}                    & \underline{45.4}                \\
YOLOX-L                                                 & \textbf{61.3}                                & 94.0                                          & \textbf{30.7}                                & 65.7                 & \textbf{34.8}         & 56.4                                         & 90.8                                          & 17.4                                         & 61.6                 & \textbf{20.2}         & 49.6                      & 71.5                                                      & \textbf{66.6}            & 17.7                     & \textbf{76.1}                & 71.7                      \\
\begin{tabular}[c]{@{}c@{}}Center-\\ Net++\end{tabular} & 54.6                                         & 92.4                                          & \textbf{23.3}                                & 61.9                 & 24.4                  & 53.6                                         & 91.1                                          & 15.0                                         & 61.0                 & 15.6                  & 34.5                      & 59.9                                                      & 62.2                     & \textbf{27.6}            & 70.8                         & 66.6                      \\
YOLOv7                                                  & 50.5                                         & 80.6                                          & -                                            & \textbf{80.9}        & -                     & 49.0                                         & 77.7                                          & -                                            & \textbf{78.1}        & -                     & 32.8                      & 69.7                                                      & 51.6                     & 11.6                     & 60.4                         & 68.1                      \\
YOLOv5n                                                 & 60.8                                         & \textbf{95.3}                                 & -                                            & \textbf{93.4}        & -                     & \textbf{64.7}                                & \textbf{93.6}                                 & -                                            & \textbf{90.3}        & -                     & 53.2                      & \textbf{78.2}                                             & \textbf{70.8}            & \textbf{33.9}            & \textbf{76.0}                & \textbf{75.8}             \\
\begin{tabular}[c]{@{}c@{}}YOLOX\\ -Tiny\end{tabular}   & 58.5                                         & 93.1                                          & 21.6                                         & 63.4                 & 23.6                  & 54.6                                         & 90.1                                          & \textbf{17.2}                                & 59.8                 & 20.0                  & 44.4                      & 68.8                                                      & 64.3                     & 20.6                     & 75.7                         & 68.2                      \\
PP-PicoDet-L                                            & \textbf{62.9}                                & \textbf{94.7}                                 & 21.3                                         & 69.2                 & \textbf{26.8}         & \textbf{61.1}                                & \textbf{93.3}                                 & 10.3                                         & 67.3                 & 14.2                  & \textbf{54.7}             & \textbf{77.2}                                             & 64.6                     & 26.7                     & 68.5                         & \textbf{74.9}             \\
\begin{tabular}[c]{@{}c@{}}NanoDet\\ -Plus\end{tabular} & 56.5                                         & 83.6                                          & \underline{3.4}                                    & 61.1                 & \underline{3.6}             & 55.8                                         & 82.1                                          & \underline{5.1}                                    & 62.6                 & \underline{5.0}             & 41.9                      & 75.7                                                      & 61.0                     & \underline{8.7}                & 72.9                         & 74.8                      \\
\begin{tabular}[c]{@{}c@{}}TPH-\\ YOLOv5\end{tabular}   & 51.2                                         & 80.3                                          & -                                            & 59.2                 & -                     & 44.2                                         & 59.9                                          & -                                            & 55.6                 & -                     & \textbf{53.4}             & 71.2                                                      & 57.5                     & 23.9                     & 73.1                         & 72.9                      \\
\begin{tabular}[c]{@{}c@{}}UAV-\\ YOLO\end{tabular}     & \underline{28.3}                                   & \underline{42.6}                                    & -                                            & 42.8                 & -                     & \underline{25.4}                                   & \underline{39.4}                                    & -                                            & \underline{39.7}           & -                     & 29.7                      & 48.4                                                      & 16.4                     & 12.5                     & 40.1                         & \underline{47.6}                \\ \hline
\end{tabular}}
\label{result_3}
\end{table}
            \begin{figure}[t] 
	            \centering
	            \includegraphics[width=1 \linewidth]{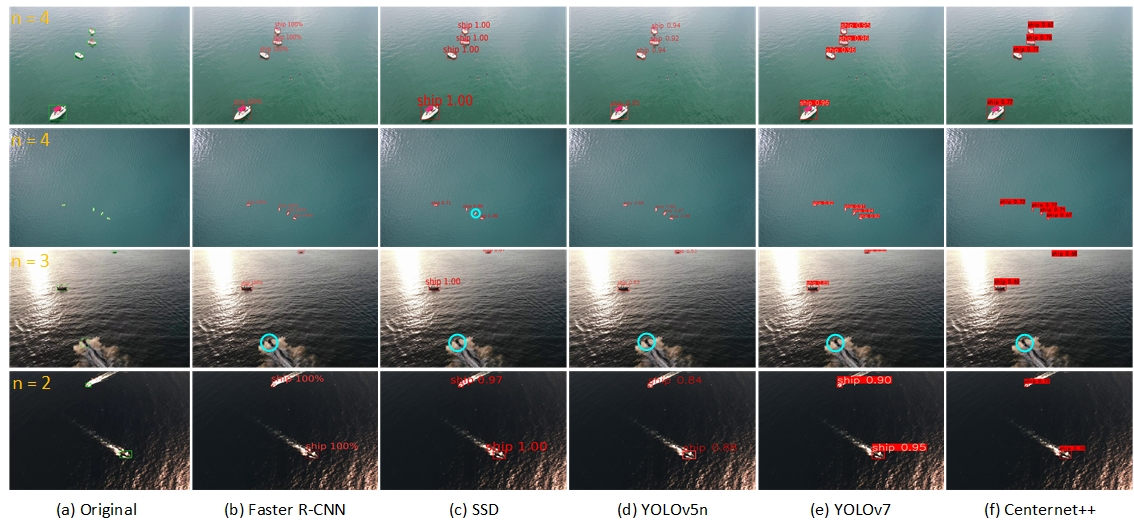}
	            \caption{The visual comparisons of several competing methods for ship detection on our MS2ship dataset. The green rectangles, red rectangles, and cyan circles correspond to the ground truth, detection results, and undetected ships, respectively. In the first column, the top left 'n =' of images means the number of ships, which is the same as the following visual images.}
	            \label{visual_result_1} 
            \end{figure}
            \begin{figure}[t]
	            \centering
	            \includegraphics[width=1 \linewidth]{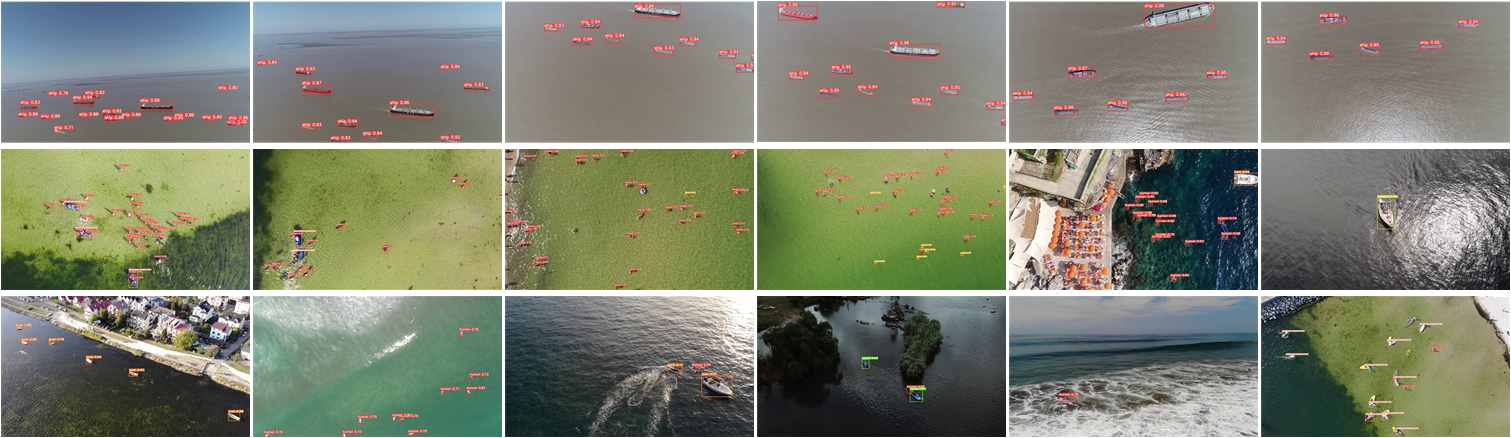}
	            \caption{The visual results for maritime object detection on the ShipDataset and AFO dataset, corresponding to the first and other two rows of the diagram.}
	            \label{visual_result_2} 
            \end{figure}
            \textbf{Detection Accuracy.} From Table \ref{result_1}, we find that YOLOv7 achieves the best performance in generic methods with 82.2$\%$ AP, 99.5$\%$ $AP_{50}$ and 99.0$\%$ AR in the trainval set. It also gets comparative performance in the test set. The YOLOv5n and TPH-YOLOv5 tie for the second place in ranking with an overall AP of 82.8$\%$ in the trainval set. In the test set, YOLOv5n still maintains good performance, but TPH-YOLOv5 reduces in all metrics. It means that the improved TPH-YOLOv5 is easy to overfit in the trainval dataset. For SOD, both Faster R-CNN and CenterNet++ acquire the top second place in the ranking on the trainval dataset. Faster R-CNN attains high performance with an $AP_{50}$ of 99.0$\%$, whereas it underperforms in the test set. Noting that the CenterNet++, an anchor-free detector, still performs best for SOD, with 65.4$\%$ $AP_S$ and 69.4$\%$ $AR_S$ in the test set, we can find that anchor-free object detection methods are good at learning small object features. As a specially designed lightweight object detector, Nanodet gets the poorest performance among all detectors in both trainval and test sets. 
            We can find that the input size of Nanodet is just 416×416, which may be one of the reasons to explain its bad performance. The input size of YOLOX-Tiny is the same as Nanodet, and the performance of YOLOX-Tiny is also poor. For the two detectors designed for UAV aerial images, they have the average performance on the MS2ship dataset. Moreover, with the largest input size of 1280×1280, UAV-YOLO has a poorer performance than TPH-YOLOv5 with an input size of 640×640. 
            For the results on the ShipDataset, as shown in Table \ref{result_2}, YOLOv5n and PP-PicoDet-L achieve the best performance, while YOLOX-L and YOLOX-Tiny get the worst detection result. Note that CenterNet++ and PP-PicoDet-L work best for SOD. For the results on the AFO dataset, as shown in Table \ref{result_3}, YOLOv5n and PP-PicoDet-L also get the best performance on mAP in the test dataset. RetinaNet and UAV-YOLO get the worst performance. Moreover, we provide the AP of every category. The detection effect is largely related to the number of object distances. Thus, we conclude that the sample imbalance of maritime objects is very significant, which is the main reason for the large difference in their detection accuracy.
            For the sake of better understanding, some of the visual comparisons over several competing methods are illustrated in Fig. \ref{visual_result_1} and Fig. \ref{visual_result_2}.

\begin{figure}[t]
        	   \centering
        	   \includegraphics[width=0.7 \linewidth]{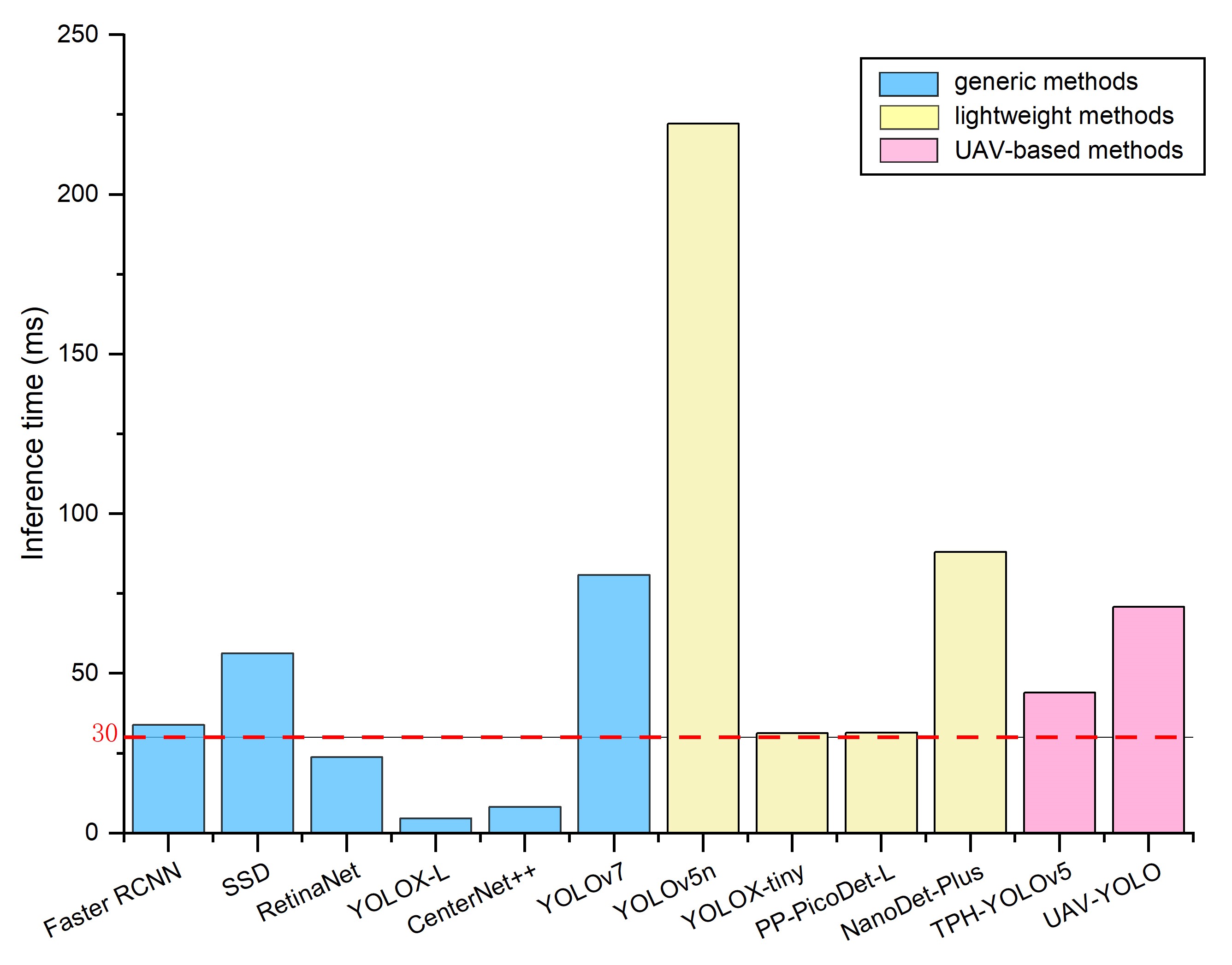}
        	   \caption{The inference time of all object detection methods in our experiment. We choose 30 FPS as the base level that could achieve real-time detection, because most videos are captured at close to 30 frame rate. Most models can detect objects in real-time, and the YOLOv5n achieves the highest detection speed.}
        	   \label{FPS} 
            \end{figure}

            \textbf{Detection speed and model size.} The model size and inference speed of the same method after training on different datasets are the same. Therefore, we only analyze the detection speed and model size according to the detection results of MS2ship dataset in Table \ref{result_1}. Besides the four lightweight models, the size of other models is on the same order of magnitude, with Params ranging from 20M to 50M and FLOPS around 100G. On the GPU with sufficient computing resources, most of the object detectors have processing speed faster than 30 FPS, as shown in Fig. \ref{FPS}. However, the YOLOX-L has the most Params with a value of 54.15M, and it has the lowest inference speed with 4.65 FPS. The FPS of Centernet++ is also lower than 10, and they cannot achieve real-time detection. The TPH-YOLOv5 has the most FLOPS of a value of 259.6G. Note that the YOLOv5n has the least Params (1.76M) and the second least calculations (4.2 GFLOPS), and achieves the fastest speed (222.22 FPS) and the highest mAP (82.8$\%$), which strikes a good balance between Params, detection accuracy, and speed, satisfying the practical applications for maritime UAVs. To sum up, YOLOv5n has high detection accuracy and the fastest detection speed with a small model size, which is the best detector on the MS2ship dataset. It also gives a suitable choice for UAV-based object detection tasks. For the common small objects in maritime UAV aerial images, the anchor-free Centernet++ has great potential due to the best performance on SOD. Because the current detectors designed for UAV aerial images, such as TPH-YOLOv5 and UAV-YOLO, are usually improved based on the Visdrone dataset, they just have normal performance on maritime object detection.

        \subsubsection{Robustness Analysis}
            The influence of haze and low-light conditions puts forward high requirements for the robustness of object detection. By adding haze and low-light conditions to some images in the test set of MS2ship dataset, we evaluate the inference models under unfavorable environments. The hazy images are acquired by the classic atmospheric scattering model, and those images in the low-light condition are handled by gamma transform. In order to maintain the authenticity of processed images, we carefully chose and processed the images without clear flares. The results in Table \ref{result_4} show the comparison of robustness between object detection methods.

\begin{table}[]
	\renewcommand\arraystretch{1.2}
	\caption{\centering Robust comparisons of object detection methods on the MS2ship dataset. The \textbf{bold} indicates that the value of metrics ranks in the top 2 of all detectors, and the \underline{italics} indicates that the value ranks at the bottom.}
	\footnotesize
	\begin{tabular}{ccccccccccc}
		\hline
		& \multicolumn{5}{c}{Test   (hazy)}                                                                                                                                                                                                               & \multicolumn{5}{c}{Test   (low-light)}                                                                                                                                                                                                          \\ \cline{2-11} 
		&                                     &                                       &                                      &                               &                                &                                     &                                       &                                      &                               &                                \\
		\multirow{-3}{*}{Methods} & \multirow{-2}{*}{AP} & \multirow{-2}{*}{$AP_{50}$} & \multirow{-2}{*}{$AP_S$} & \multirow{-2}{*}{AR} & \multirow{-2}{*}{$AR_S$} & \multirow{-2}{*}{AP} & \multirow{-2}{*}{$AP_{50}$} & \multirow{-2}{*}{$AP_S$} & \multirow{-2}{*}{AR} & \multirow{-2}{*}{$AR_S$} \\ \hline
		Faster RCNN-FPN                      & 73.3                                                       & 95.7                                                         & 59.8                                                        & 77.0                            & 64.8                           & 74.1                                                       & 95.8                                                         & 58.0                                                          & 77.6                          & 62.9                           \\
		SSD                                  & 72.2                                                       & 96.5                                                         & 54.1                                                        & 76.5                          & 61.7                           & 72.7                                                       & 96.5                                                         & 54.7                                                        & 77.0                            & 62.6                           \\
		RetinaNet                            & 73.7                                                       & \textbf{97.1}                                                & 56.8                                                        & 78.9                          & 65.7                           & 74.3                                                       & 96.3                                                         & 56.7                                                        & 79.1                          & 64.8                           \\
		YOLOX-L                              & 75.7                                                       & 96.8                                                         & \textbf{62.3}                                               & 78.7                          & \textbf{67.1}                  & 76.1                                                       & 96.8                                                         & \textbf{62.1}                                               & 79.1                          & \textbf{66.6}                  \\
		CenterNet++                          & 72.0                                                         & 95.7                                                         & \textbf{62.1}                                               & 76.7                          & \textbf{66.2}                  & 72.5                                                       & 95.8                                                         & \textbf{61.4}                                               & 77.3                          & \textbf{65.4}                  \\
		YOLOv7                               & 79.8                                              & \textbf{97.6}                                                & -                                                           & \textbf{97.2}                 & -                              & 80.1                                                       & \textbf{97.8}                                                & -                                                           & \textbf{97.0}                   & -                              \\
		YOLOv5n                              & \textbf{80.0}                                                         & 96.9                                                         & -                                                           & \textbf{94.1}                 & -                              & \textbf{81.0}                                                & 97.2                                                         & -                                                           & \textbf{94.6}                 & -                              \\
		YOLOX-Tiny                           & 68.4                                                       & 93.8                                                         & 53.4                                                        & 71.6                          & 57.8                           & 70.2                                                       & 95.8                                                         & 55.8                                                        & 73.3                          & 60.2                           \\
		PP-PicoDet-L                         & 73.6                                                       & 96.5                                                         & 56.0                                                          & 77.2                          & 62.1                           & 76.5                                                       & \textbf{97.5}                                                & 59.4                                                        & 79.7                          & 65.2                           \\
		NanoDet-Plus                         & \underline{59.6}                                        & \underline{87.5}                                          &  \underline{32.4}                                         & \underline{64.0}             & \underline{37.4}            &  \underline{61.9}                                       & \underline{88.9}                                          &  \underline{33.7}                                         & \underline{65.9}           & \underline{39.1}          \\
		TPH-YOLOv5                           & \textbf{81.0}                                                & 95.6                                                         & -                                                           & 91.7                          & -                              & \textbf{81.2}                                              & 95.8                                                         & -                                                           & 92.2                          & -                              \\
		UAV-YOLO                             & 70.4                                                       & 93.2                                                         & -                                                           & 91.9                          & -                              & 70.9                                                       & 93.7                                                         & -                                                           & 93.6                          & -                              \\ \hline
	\end{tabular}
	\label{result_4}
\end{table}
            \begin{figure}[t]
	            \centering
	            \includegraphics[width=1 \linewidth]{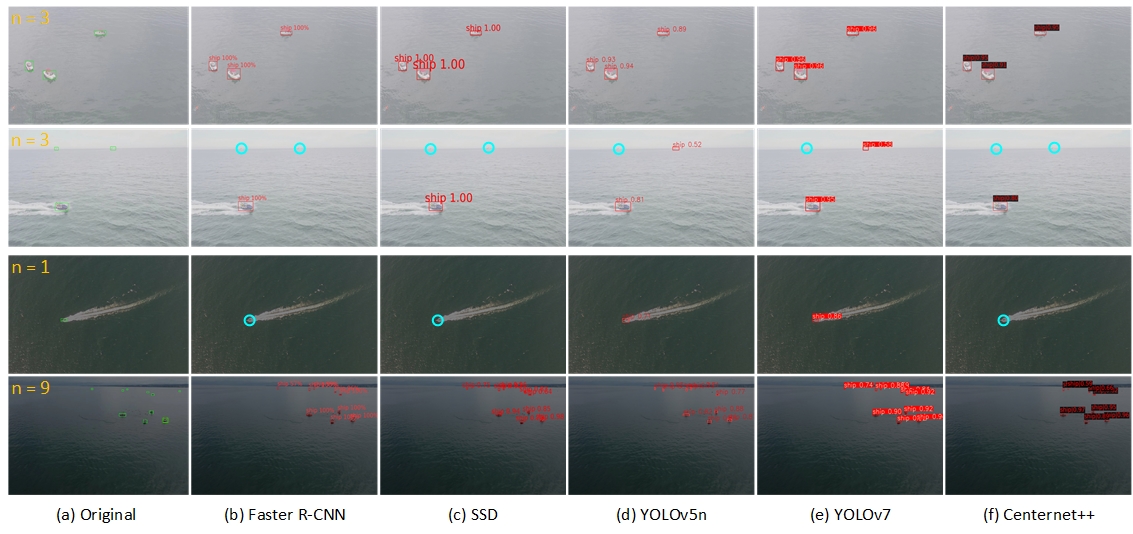}
	            \caption{The visual comparisons of several competing methods for ship detection under hazy and low-light conditions, corresponding to the top and bottom half of the diagram. The green rectangles, red rectangles, and cyan circles correspond to the ground truth, detection results, and undetected ships, respectively.}
	            \label{visual_result_3} 
            \end{figure}
            In general, under haze and low-light conditions, the performance of all detectors declines to varying degrees. For Faster R-CNN and CenterNet++, we can find that the AP under the haze condition is lower than that under the low-light condition, but the $AP_S$ is the other way around. It explains that the influence of the low-light condition on SOD is serious for these two detectors. Under the haze condition, the AP of Faster R-CNN, CenterNet++, YOLOX-Tiny, PP-PicoDet-L, and NanoDet-Plus reduced by more than 3$\%$, meaning that the haze highly influenced these detectors. According to the result, the AP of YOLOX-Tiny decreased by 6.4$\%$. Due to the declining AP bigger than 3$\%$, the impact of the low-light condition is great for Faster R-CNN, CenterNet++, and YOLOX-Tiny. The YOLOX-Tiny is also the most affected detector with 4.6 percentage points falling. For PP-PicoDet-L and NanoDet-Plus among lightweight methods, the low-light influence is small with AP decreases of 0.5$\%$ and 0.7$\%$, respectively. However, these two detectors are susceptible to the haze. About Faster R-CNN and CenterNet++ for SOD, due to the falling $AP_S$ higher than 3$\%$, the effects of haze and low-light conditions are both enormous. It is worth noting that YOLOX-Tiny and PP-PicoDet-L are easy to be affected by haze, while the effect of the low-light condition is faint. As a whole, the detection performance of YOLOv7 and YOLOv5n is the best, whether there are haze and dim lights or not. 
            For the sake of better understanding, some of the visual comparisons under haze and lowlight conditions are illustrated in Fig. \ref{visual_result_3}.

    \subsection{Discussion}
        According to the benchmarking and robustness analysis, we find that there are some limitations for maritime object detection in maritime UAV aerial images, which are summarized as follows.

        Firstly, the cover area of foreground objects on UAV aerial images is fewer than that on natural images. The huge difference in cover areas between foreground and background objects will cause the problem of the imbalance between positive and negative samples. The learning progress is thus easy to be disturbed by background noises. The intense background noises caused by complicated maritime scenes will result in a bad performance on UAV aerial images. 
        By focusing on the interested objects and weakening background noises, the attention mechanism could effectively deal with the background noise interference problem \citep{liu2020small, hou2021novel}.

        Secondly, we adopt pre-trained backbone models in all object detection methods, which leads to limitations in the specific task. The pre-trained models contain a huge number of Params, and have limited structure design space. The flexibility to control and adjust the network structure is just restricted, even if there are small changes. 
        Therefore, it is better to train the model from scratch for maritime object detection. Moreover, in the same scenario captured by maritime UAV, the detection results of different objects are still different. It is mainly caused by the imbalanced instance distribution. Different network structures also lead to disparate attention on various features.

        Thirdly, there has not been a large-scale benchmarking dataset for maritime UAV-based object detection yet, and the existing studies seldom use the same maritime UAV aerial dataset. In some cases, the simulated data are considered to support the specific study. 
        \cite{alvey2021simulated} combined the photorealistic simulation, open source libraries, and high-quality content to develop a workflow to assist the CV research for UAVs. The Maritime Synthetic (MarSyn) dataset \citep{ribeiro2022real} consists of 25 different video sequences captured by analog cameras from the UAV perspective. These synthetic data overcome the difficulty of obtaining data in real scenarios.

        Finally, when it comes to the practical deployment of detection models, it is necessary for maritime UAVs to improve the ability to monitor sea-surface in real-time. However, the lightweight design of models will inevitably sacrifice some detection accuracy, which has been discussed in the lightweight methods before. It still needs much effort to reach a fine trade-off between detection accuracy and computational cost.

        To promote the visual applications of maritime UAVs, it is necessary to continue the research on lightweight detection methods for small objects. We also have observed several trends of maritime UAVs that will be potentially developed in the future, which can be listed in 5 parts.

        \textbf{Image Preprocessing Technology.} The imaging quality for cameras loaded on maritime UAVs is always unfriendly due to the greatly changeable maritime environments. The sharp motion of the camera and observed objects will bring noises and blurs to the videos/images. In addition, the poor visibility, under haze and low-light conditions, has a large impact on object detection. Therefore, it is helpful to improve the image quality by proper preprocessing operations, i.e., image denoising \citep{kong2020comprehensive}, deblurring \citep{zhang2022deep}, dehazing \citep{gui2021comprehensive} and enhancement \citep{cheng2023joint}. For instance, \cite{zheng2023uav} proposed a saliency-guided parallel learning mechanism to remove UAV aerial image haze. It takes the particular imaging mechanism of UAVs into consideration. In order to tackle the artifacts caused by low resolution, motion blur, and other environmental factors, \cite{agarwal2021impact} studied state-of-the-art super-resolution algorithms and improved the quality of aerial images.

        \textbf{Data Augmentation Technology.} A deep learning-based object detection model relies on large quantities of training samples. However, due to the relatively high acquisition cost of maritime UAV aerial images, dataset scarcity is still a trouble. The data augmentation techniques assist in increasing the size and scale of training datasets, which is always used to relieve the scarcity problem. An adaptive resampling augmentation strategy can logically augment the UAV aerial images \citep{chen2019rrnet}. \cite{shin2020data} extracted the mask of foreground objects and combined them with new backgrounds, automatically generating the location and classification information. In some special environments, there may be a prohibition by laws. Therefore, \cite{kiefer2022leveraging} discovered the potential application of synthetic data and expands the open-source framework, named DeepGTAV, to apply to the UAV aerial scenarios. \cite{yu2021pp} indicated that excessive data augmentation can heighten the regularization effect and make the training difficult to converge. It is an interesting direction to explore the data augmentation strategy that is suitable for maritime UAV aerial images.

        \textbf{Semantic Segmentation Technology.} The semantic segmentation for UAV remote sensing images is one of the emerging and challenging research focuses at present. Compared with the satellite-based remote sensing platform, the UAV platform can fly close to the sea surface to improve the resolution of maritime objects. Through assigning a class label to each pixel in the UAV aerial image, semantic segmentation assists to facilitate maritime scene understanding. However, this advanced visual task also encounter the problem of insufficient computing and storage capacity. A survey on real-time semantic segmentation manifests that compact and efficient models can work well on resource-constrained hardware and keep balance between latency and accuracy \citep{holder2022efficient}. \cite{liu2021light} proposed an efficient lightweight network with a few parameters by taking the EfficientNet-b1 \citep{tan2020efficientdet} as a backbone, as well as the feature extraction structure designed by the NAS approach. The recently popular segment anything model (SAM) \citep{kirillov2023segment} has greatly promoted its applications in various domains, and it is a worthy direction to look forward to the introduction of SAM on maritime UAVs.  

        \textbf{Multi-sensor Fusion Technology.} The maritime UAVs are capable of obtaining enough comprehensive and accurate information by multi-sensor fusion technology. Using a single sensor results in the information inconsistency and insufficient accuracy. To achieve easier and faster maritime object detection of maritime UAVs under complex environments, it is an efficient way to combine and fuse images from visible and infrared sensors \citep{jiang2017uav}. The cameras just provide visual information, which is easy to be affected by the harsh maritime environments. A multi-modal sensor fusion approach was proposed by \cite{stanislas2018multimodal} to improve the robustness of detection and classification performance. \cite{wu2022new} proposed a multi-sensor fusion perception system, which was able to improve the information consistency and data accuracy of ship motion. Moreover, ship trajectory prediction technology \citep{zhang2023deep} is also an important part for maritime data fusion. However, most of the existing multi-source data fusion methods utilize multi-sensors equipped on ships or shore \citep{9731523, qu2023improving}. There is a potential and start-up direction to realize the data fusion by using maritime UAV-based cameras and other sensors, which is also beneficial for detection and localization accuracy \citep{zhou2019verification, xiu2019multi, kim2020operational}.

        \textbf{Task Offloading Strategy.} The computation-intensive tasks usually require for the high processing performance supported by large models. But many deep learning-based models are so cumbersome that are not suited to deploy on maritime UAV platforms. One of the effective solutions is to offload the task to on-ground servers or nearby UAVs whose computing and energy resources are available. In the edge computing scenario, the maritime UAVs offload tasks, such as object detection and semantic segmentation, to the edge server located at the shore-based or ship-borne station, and then the edge server returns processing results to the UAVs. In this way, the energy consumption and processing delay of maritime UAVs will be greatly reduced. A UAV-enabled fog computing network proposed by \cite{li2020multi} provided an effective strategy for the multi-task offloading. However, there are latency and security issues on the transmission of images and videos. 
        Considering that the video transmission bandwidth was limited, \cite{rudol2019evaluation} evaluated deep learning-based object detection methods by using common video compression technologies. There are two kinds of task offloading methods, namely traditional methods based on heuristic algorithms and intelligent methods based on online learning, and the latter is the focus of future research \citep{islam2021survey}.

\section{Conclusion and Future Perspectives}   
    In this paper, we surveyed more than 200 recent studies about UAV-based object detection by using deep learning technologies. We first discussed the specific challenges for object detection on maritime UAVs, and presented a systematic study on the detection methods for UAV aerial images. Concretely, about the challenge of object feature diversity, we provided an overview of scale-aware, view-aware, and small object detection methods. Regarding the poor efficiency issue caused by device limitation, we reviewed the lightweight methods. We also presented a review of rotated object detection and other methods. Moreover, we reviewed UAV aerial image/video datasets all along with their characteristics, and then proposed our MS2ship dataset for ship detection on maritime UAVs. In addition, we presented evaluation metrics and analyzed experimental results by using state-of-the-art object detection methods. Lastly, we presented the discussion of limitations, interrelated and potential works about the intelligent development of maritime UAVs. By the image preprocessing and data augmentation techniques, the detection accuracy will obtain a certain improvement, but it is still a critical challenge to speed the inference process in the model deployment stage. The task offloading is an effective solution to deal with the deployment of large models, which is suitable for some specific applications. To make maritime UAVs more intelligent and practical, future work will focus on the semantic segmentation and multi-source data fusion technologies.


%
\section*{Declaration of competing interest}
The authors declare that they have no known competing financial interests or personal relationships that could have appeared to influence the work reported in this paper.
\section*{Acknowledgments}
The work described in this paper was supported by grants from the National Key Research and Development Program of China (No.: 2022YFB4300300), and the Research Project of Wuhan University of Technology ChongQing Research Institute (No.: YF2021-13).
\bibliographystyle{cas-model2-names}

\bibliography{cas-refs}
%
%
%
\end{document}